\PassOptionsToPackage{table,dvipsnames}{xcolor}
\documentclass{fairmeta}
\usepackage{amsmath,amssymb,latexsym}
\usepackage{array,makecell,tabularx,colortbl,adjustbox}
\usepackage{algorithm,algpseudocode}
\usepackage{pifont,textcomp,wasysym}
\usepackage{enumitem,xurl,needspace}
\newcolumntype{Y}{>{\raggedright\arraybackslash}X}
\newcommand{\cmark}{\ding{51}}
\newcommand{\xmark}{\ding{55}}

\ifXeTeX
  \DeclareFontShape{TU}{\sfdefault}{b}{sc}{<->ssub * \sfdefault/b/n}{}
  \DeclareFontShape{TU}{\sfdefault}{bx}{sc}{<->ssub * \sfdefault/b/n}{}
\fi


\title{PixSearch: Region-Grounded Retrieval for Knowledgeable Large Multimodal Models}

\author[1,2,*]{Jeonghwan Kim}
\author[1]{Renjie Tao}
\author[1]{Sanat Sharma}
\author[1]{Jiaqi Wang}
\author[1]{Kai Sun}
\author[1]{Zhaojiang Lin}
\author[1]{Seungwhan Moon}
\author[1]{Lambert Mathias}
\author[1]{Anuj Kumar}
\author[2]{Heng Ji}
\author[1]{Xin Luna Dong}

\affiliation[1]{Meta Reality Labs}
\affiliation[2]{University of Illinois Urbana-Champaign}

\contribution[*]{Work done at Meta}
\abstract{
Multimodal Retrieval-Augmented Generation (MM-RAG) augments visual question answering (VQA) with external knowledge, but many systems rely on modular pipelines that separate visual grounding, query construction, and answer generation. This separation can propagate errors from intermediate detections, masks, or captions into retrieval while adding latency and limiting coordination with ongoing reasoning. These limitations are especially consequential for egocentric images, where questions often concern small, off-center, or long-tail entities that global retrieval can overlook. We propose \textsc{PixSearch}, an end-to-end segmenting Large Multimodal Model (LMM) that unifies region-level visual grounding with retrieval-augmented reasoning. During autoregressive generation, \textsc{PixSearch} learns when to retrieve, how to route queries across text, whole-image, and region-level retrieval, and how to crop the target entity as a visual query using a jointly trained mask decoder. A two-stage supervised fine-tuning regimen with search-interleaved supervision teaches retrieval triggering, query selection, and evidence-grounded answer generation while preserving segmentation ability. On egocentric and entity-centric VQA benchmarks, \textsc{PixSearch} achieves an 18.3\% relative accuracy gain on CRAG-MM over whole-image retrieval, with particularly strong gains on egocentric images, while remaining competitive on broader VQA and text-only QA tasks. It also reduces inference latency by 36.4\% compared with a cached GroundedSAM--LLaVA pipeline.
}

\date{October 12, 2026}
\correspondence{Jeonghwan Kim, \email{jk100@illinois.edu}; \email{jeonghkim@meta.com}}
\metadata[Code]{\url{https://github.com/wjdghks950/PixSearch}}
\hypersetup{
  pdftitle={PixSearch: Region-Grounded Retrieval for Knowledgeable Large Multimodal Models},
  pdfauthor={Jeonghwan Kim, Renjie Tao, Sanat Sharma, Jiaqi Wang, Kai Sun, Zhaojiang Lin, Seungwhan Moon, Lambert Mathias, Anuj Kumar, Heng Ji, Xin Luna Dong},
  pdfsubject={Region-grounded multimodal retrieval-augmented reasoning},
  pdfkeywords={PixSearch, multimodal retrieval, visual question answering, segmentation}
}
\begin{document}
\maketitle
\section{Introduction}
\begin{figure}[!htbp]

    \centering
\begin{minipage}{.72\textwidth}
\centering
    \includegraphics[width=\linewidth]{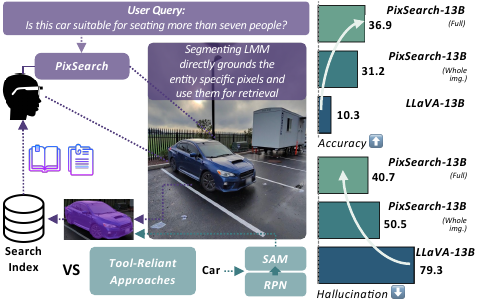}

    \caption{Egocentric images from wearable devices often render small entities. Full-image or caption-only MM-RAG can retrieve noisy evidence, while \textbf{\textsc{PixSearch}} learns when to search, whether to query text, image, or region, and how to reason over retrieved evidence. On CRAG-MM~\cite{wang2025crag}, \textbf{\textsc{PixSearch}} improves overall accuracy by 18.3\% and reduces hallucination by 19.4\% relative to whole-image retrieval.}
    \label{fig:smart_glass_illustration}

\end{minipage}
\end{figure}

Wearable devices such as smart glasses are emerging as always-on, hands-free assistants that see the world from the user's viewpoint. Realizing this vision requires \emph{entity-centric visual question answering} (VQA): recognizing the specific entities in front of the user, retrieving factual knowledge about them, and composing evidence to answer situated questions. For example, as illustrated in Figure~\ref{fig:smart_glass_illustration}, deciding whether a visible car can fit seven people requires knowledge about that specific vehicle, not merely recognizing that a car is present. Yet egocentric VQA is challenging: wide-angle cameras often make entities small, off-center, or ambiguous, and many everyday objects are long-tail entities unlikely to be reliably covered by an Large Multimodal Model's (LMM) internal knowledge.

Multimodal Retrieval-Augmented Generation (MM-RAG) has strengthened factual grounding in VQA \citep{marino2021krisp, lin2022revive, jian2024llmra}, but two limitations persist. First, most MM-RAG approaches either retrieve with the full image \citep{shah2019kvqa, marino2021krisp, yang2023re, yan2024echosight, yu2025visrag, ha2025mm, sidhu2025search}, or use text-only queries that are simple captions of the image~\citep{narasimhan2018straight, garderes2020conceptbert, gao2022transform, salaberria2023imgcapretrieve}. 
Full-image retrieval pulls in distracting background, while text-only cues (e.g., ``car'') lack the specificity needed for fine-grained entity grounding and enrichment. Second, MM-RAG pipelines are often modular---detectors, segmenters, captioners, etc.---to form queries, thus can introduce cross-modal translation errors, struggle with composing multiple queries, and add latency when retrieval is \textit{not} truly necessary.

We present \textsc{PixSearch}, the first \textit{end-to-end} framework for retrieval-augmented reasoning. During generation, \textsc{PixSearch} (i) learns \emph{when} to retrieve by emitting \texttt{<search>} tokens, (ii) decides \emph{how} to retrieve by routing among text, whole-image, and region-level queries via token outputs, and (iii) grounds answers in the retrieved evidence, supporting multi-step search. Built on segmenting LMMs (Large Multi-modal Models with segmentation capabilities), \textsc{PixSearch} produces segmentation masks end-to-end without external detection/segmentation APIs, and uses these masks directly as retrieval queries. This yields pixel-level, context-aware grounding that surpasses modular, text- or tool-driven pipelines.

Training \textsc{PixSearch} requires learning retrieval triggering, query-type selection, region grounding, and evidence-based reasoning. We introduce a two-stage supervised fine-tuning strategy (\S\ref{sec:two_stage_training}) and a search-interleaved data construction pipeline (Figure~\ref{fig:data_construction}) to teach these behaviors while preserving segmentation ability. We also compare joint SFT and Search-R1-style reinforcement learning~\citep{jin2025searchr1} to assess the trade-off between answering accuracy and grounding retention.

In summary, our paper makes the following three contributions.
\begin{enumerate}[leftmargin=*, itemsep=1pt, topsep=2pt, parsep=0pt, partopsep=0pt]
    \item {\em Model \& Framework:} We introduce {\sc PixSearch}, the first end-to-end segmenting LMM that autonomously triggers retrieval and performs region-level grounding.
    \item {\em Training}: We devise a two-stage supervised fine-tuning regimen and a multimodal training dataset that teaches when to retrieve and how to form text, image, or region queries via search-interleaved trajectories, while preserving segmentation quality.
    \item {\em Experiments}: Through comprehensive experimental results, we demonstrate that \textsc{PixSearch} substantially improves factual consistency and generalization across egocentric and entity-centric VQA benchmarks, achieving 18.3\% relative accuracy improvement over whole-image retrieval in CRAG-MM~\cite{wang2025crag}, and in particular 22.3\% improvement on egocentric images (Figure~\ref{fig:smart_glass_illustration}).
\end{enumerate}

\section{Related Work}
\label{sec:related_work}

\paragraph{Knowledge-based Visual Question Answering.}
Prior KB-VQA work shows that external knowledge improves factual grounding beyond parametric model knowledge. Methods such as KRISP~\citep{marino2021krisp}, KAT~\citep{gui2022kat}, and MAVEx~\citep{wu2022mavex} retrieve symbolic or textual evidence for detected objects or candidate answers, demonstrating the value of retrieval-augmented reasoning. However, these methods largely query with text-based entity representations, such as object tags or noun phrases. More recent approaches use detector outputs as region-specific queries~\citep{lin2022revive} or LLM-extracted referring expressions for visual grounding~\citep{jian2024llmra}, highlighting the promise of region-centric retrieval. Yet they still rely on external detectors, segmentation modules, or hand-engineered pipelines, whereas \textsc{PixSearch} learns region-level retrieval end-to-end.

\paragraph{Multimodal Retrieval-Augmented Generation.}
RAG has also been extended to multimodal domains following their effectiveness in the text-only domain. EchoSight~\cite{yan2024echosight} combines visual search with textual retrieval for encyclopedic VQA, VisRAG~\cite{yu2025visrag} targets visually rich documents, RA-CM3~\cite{yasunaga2023retrieval} unifies text and image retrieval for generation, and UniIR/M-BEIR~\cite{wei2024uniir} benchmarks multimodal retrieval. These works ground multimodal models in external knowledge, but typically rely on text-only or whole-image queries, limiting precision when the relevant evidence concerns a specific entity in the scene.

\paragraph{Search Trigger and Interleaving External Knowledge.}
Another challenge is teaching models not only to use retrieval, but to decide when to retrieve and how to query. Self-RAG~\cite{asai2024selfrag}, Toolformer~\cite{schick2023toolformer}, and RePlug~\cite{shi2024replug} introduce adaptive retrieval or tool-use mechanisms, while Search-R1~\citep{jin2025searchr1} uses reinforcement learning to interleave search with autoregressive reasoning. These methods show that adaptive retrieval can reduce unnecessary latency and hallucinations, but they primarily operate in text-only settings and do not address multimodal region-level retrieval for fine-grained visual reasoning.

\paragraph{Segmenting Large Multimodal Models.}
Segmenting LMMs, including LISA~\cite{lai2024lisa}, PixelLM~\cite{ren2024pixellm}, GLaMM~\cite{rasheed2024glamm}, Osprey~\cite{yuan2024osprey}, and PLUM~\cite{blume-kim-2025partonomy}, generate segmentation masks alongside text by predicting mask tokens or embeddings inline. These models show that segmentation can be integrated into LMM reasoning and enable fine-grained visual grounding. However, they answer questions using only internal model knowledge; \textsc{PixSearch} extends this line by using generated regions as retrieval queries for external knowledge.

\section{Overview}
\label{sec:overview}
\subsection{Problem Definition}
We study the {\em Visual Question Answering (VQA)} problem, which takes an image $I$ and a question $Q$ regarding the image as input, and outputs an answer $A$ to the question. We assume an external knowledge repository to facilitate question answering. A good answer shall be relevant and helpful, and meanwhile consistent with knowledge present in the repository. 

We assume the knowledge repository is accessible through a retrieval API $\texttt{search\_api}(S,k)$, which returns the top-$k$ relevant text chunks on search query $S$. The query can be in three forms: (1) full image, where the API returns information about the image; (2) masked region, normally for a particular entity and the API returns information about the entity; and (3) text span, where the API returns search results for the text query. 

An effective MM-RAG solution needs to make the following decisions: 1) whether to issue (a single or multiple) search queries; 2) the modality of each query---full image, masked region, or text span; 3) the specific image region or text span to query; 4) the final answer based on the retrieval evidence. Existing pipeline-based methods separate these abilities, introducing translation errors and instability. We next describe a uniform framework that resolves all four through an interleaved search-and-generation decoding process.

\begin{figure}[!htbp]
    \centering
\begin{minipage}{.94\textwidth}
\centering
    \includegraphics[width=\linewidth]{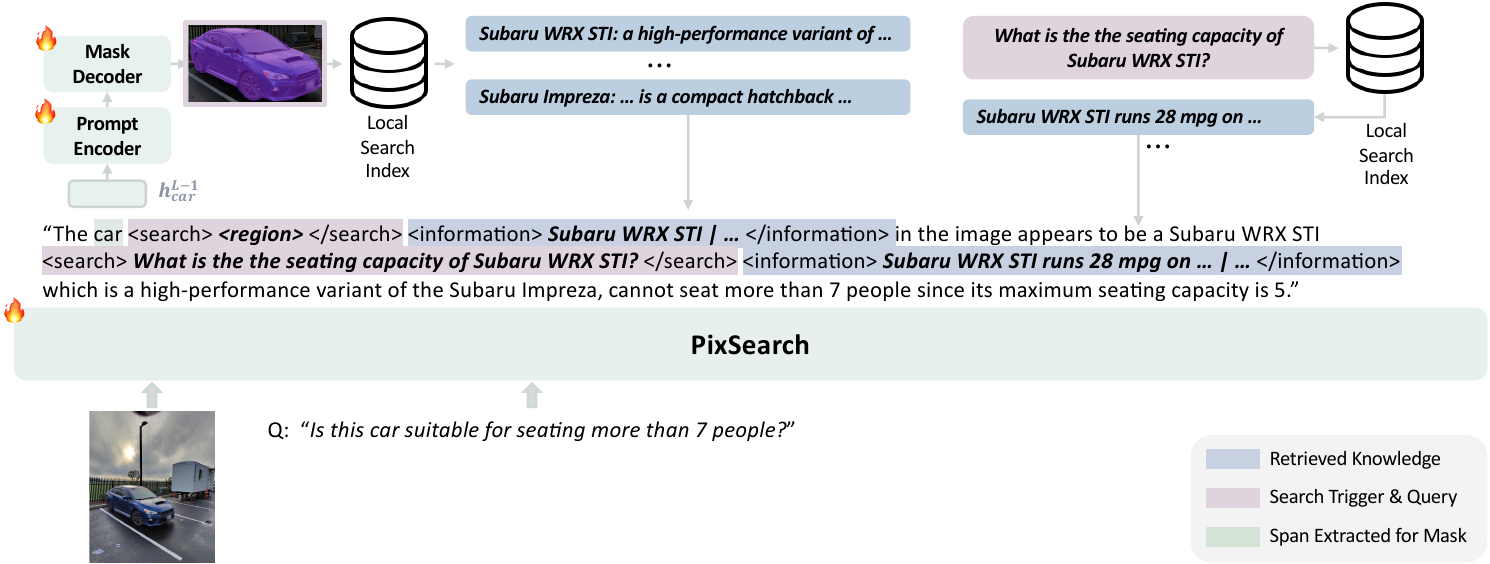}
    \caption{\textbf{Overview of the proposed \textsc{PixSearch} framework.} The model learns to decide \emph{when} retrieval is needed, \emph{how} to query (text, whole image, or segmented region), and grounds its answers in retrieved evidence while preserving mask-generation capabilities.}
    \label{fig:PixSearch_overview}
\end{minipage}
\end{figure}

\subsection{The PixSearch Framework}
\label{sec:framework}
\textsc{PixSearch} interleaves generation and retrieval (Figure~\ref{fig:PixSearch_overview}). After \texttt{<search>}, it emits \texttt{<image>}, \texttt{<region>}, or a text query $T$. Region queries use the predicted mask $\hat M=f_\theta(I,x_{<t})$:
\begin{equation}
S=\begin{cases}
I, & \text{whole-image query},\\
\texttt{crop}(I,\hat M), & \text{region query},\\
T, & \text{text query}.
\end{cases}
\end{equation}
After \texttt{</search>}, $\texttt{search\_api}(S,k)$ returns the top-$k$ text chunks, inserted between \texttt{<information>} and \texttt{</information>} (abbreviated \infstart{} and \infend{}). Generation resumes conditioned on this evidence, permitting further searches until end-of-sequence. Section~\ref{sec:formal_decode} formalizes this process.

\subsection{Formal Search-Interleaved Decoding}
\label{sec:formal_decode}

Figure~\ref{fig:PixSearch_overview} depicts the PixSearch framework. \textsc{PixSearch} conducts search-interleaved decoding, a retrieval-augmented generation process that enables the model to decide \emph{when} to retrieve and \emph{how} to ground retrieved evidence in its multimodal reasoning trajectory. At each decoding step $t$, the model autoregressively predicts the next token $x_t$ based on the image $I$ and the previously generated tokens, until an end-of-sequence (\texttt{</s>}) token is reached:  
\begin{equation}
\label{eqn:decode}
x_t \sim p_\theta(x_t \mid x_{<t}, Q, I).   
\end{equation}

An output token can be a special control token \texttt{<search>}, at which point the model temporarily halts textual decoding to initiate a retrieval subroutine, which proceeds in three steps. First, the subroutine generates a \textit{payload} string in $\{\texttt{<image>}, \texttt{<region>}, \texttt{<text>}\footnote{The model directly generates the textual query instead of \texttt{<text>}.}\}$ to describe the retrieval modality. We then generate the search query for different modalities. For the \textit{image} mode, the whole input image $I$ serves as the query. For the \textit{region} mode, the model invokes its aligned mask decoder to predict a binary mask $\hat M = f_\theta(I, x_{<t})$, from which a cropped visual query is extracted. For the \textit{text} mode, the model generates the textual query during decoding. 
\begin{equation}
S =
\begin{cases}
I, & \text{if } payload=\texttt{<image>} \\
\texttt{crop}(I, \hat M), & \text{if } payload=\texttt{<region>} \\
\texttt{Text}, & \text{if } payload=\texttt{Text}
\end{cases}
\end{equation}
\begin{equation}
x_{1:t} \;\gets\; [\,x_{1:t-1}, \texttt{<search>}, S, \texttt{</search>}].
\end{equation}
Finally, the subroutine obtains the retrieved evidence:
\begin{equation}
\textit{E} = \texttt{search\_api}(S,k)
\end{equation}
where $\texttt{search\_api}$ returns a piece of textual knowledge. 
This retrieved content is then injected back into the generation stream as an \texttt{<information>} block (abbreviated as \texttt{<info>} hereafter for brevity):
\begin{equation}
\label{eqn:info}
x_{1:t+1} \;\gets\; [\,x_{1:t}, \infstart, \textit{E}, \infend\,]
\end{equation}
allowing the model to continue decoding while conditioning on the newly appended evidence.  

Through generating multiple \texttt{<search>} blocks and populating back \texttt{<information>} evidence blocks, this decoding strategy results in a dynamic reasoning trajectory that alternates between internal generation and external retrieval, enabling the model to ground answers in factual evidence whereas maintaining fine-grained visual reasoning through region-level queries.

Appendix~\ref{sec:detailed_decode} gives the algorithm walkthrough and pseudocode.

\section{\textsc{PixSearch}: Region-level Retrieval for LMMs}
\label{sec:training}

\subsection{Overview of model training strategy}
\label{sec:training:overview}
\begingroup\sloppy
At the core of \textsc{PixSearch} is a segmentation-capable Large Multimodal Model (i.e., {\em segmenting LMMs}~\citep{rasheed2024glamm, lai2024lisa, ren2024pixellm, wang2024segllm, blume-kim-2025partonomy}), with two key capabilities required by the framework: segmentation (to facilitate mask generation $f_\theta$) and search-interleaved decoding (\S\ref{sec:framework}). This design exploits the rich textual semantics for pixel-level grounding, encompassing regular open-vocabulary segmentation to referring expression segmentation, while avoiding specialized region-proposal networks \citep{ren2024grounded} or external segmentation models \citep{kirillov2023segment} (\S~\ref{sec:training:backbone}).\par\endgroup

Jointly learning segmentation and retrieval-augmented generation can degrade grounding, particularly with insufficient segmentation supervision (\S\ref{sec:mask_segmentation}). We use two stages: Stage 1 preserves segmentation and visual grounding, and Stage 2 teaches retrieval triggering, query construction, and evidence-conditioned reasoning (\S\ref{sec:training:implementation}). Table~\ref{tab:stage1_ablation} compares this schedule with joint SFT and a GRPO control.

\subsection{\textsc{PixSearch} Model}
\label{sec:training:backbone}
\paragraph{Backbone and auxiliary losses.}
We retain PLUM's text-aligned segmentation backbone and auxiliary objectives~\citep{blume-kim-2025partonomy}, adapting the language-model loss for retrieved evidence:
\begin{equation}
\mathcal{L} = \underbrace{\mathcal{L}_{\mathrm{LM}}+\lambda_{1}\mathcal{L}_{\mathrm{span}}}_{\text{sequence loss}}
+\underbrace{\lambda_{2}\mathcal{L}_{\mathrm{seg}}+\lambda_{3}\mathcal{L}_{\mathrm{BCE}}}_{\text{segmentation loss}}
+\underbrace{\lambda_{4}\mathcal{L}_{\mathrm{KL}}}_{\text{regularization}}.
\end{equation}
A bidirectional span extractor predicts BIO tags~\citep{ramshaw1999text} from final-layer token embeddings $h_i^L\in\mathbb{R}^d$, supervised by cross-entropy $\mathcal{L}_{\mathrm{span}}$. Contiguous B--I chains form entity spans $\mathcal{S}=\{(i_s,j_s)\}_{s=1}^{N_+}$. A learned projection $q_k=g(h_k^L)\in\mathbb{R}^m$ supplies mask queries to the decoder, which predicts $\hat M_i$. Mask supervision combines binary cross-entropy $\mathcal{L}_{\mathrm{BCE}}$ with Focal-Tversky loss~\citep{abraham2019novel}:
\begin{equation}
\mathcal{L}_{\mathrm{seg}}=\frac{1}{N_+}\sum_{y_i\neq\textsc{O}}\mathcal{L}_{\mathrm{FT}}(M_i,\hat M_i).
\end{equation}
A Gaussian KL constraint preserves alignment with frozen teacher embeddings $t^L_{i_s:j_s}$:
\begin{equation}
\mathcal{L}_{\mathrm{KL}}=\frac{1}{N_+}\sum_{s=1}^{N_+}
\frac{\|h^L_{i_s:j_s}-t^L_{i_s:j_s}\|_2^2}{2\sigma^2}.
\end{equation}

\paragraph{Information token masking.}
Retrieved evidence remains in context but is excluded from token supervision~\citep{jin2025searchr1}. For a length-$L$ sequence,
\begin{equation}
m_i=\begin{cases}
0, & x_i\text{ belongs to an information span},\\
1, & \text{otherwise},
\end{cases}
\end{equation}
where information spans are delimited by \texttt{<information>} and \texttt{</information>}. The masked next-token objective is
\begin{equation}
\mathcal{L}_{\mathrm{LM}}=-\frac{1}{\sum_i m_i}
\sum_{i=1}^{L}m_i\log p_\theta(y_i\mid y_{<i},I),
\label{eqn:decode-mask}
\end{equation}
where $y_i$ is the target token. This trains evidence-conditioned reasoning without supervising prediction of retrieved content.

\subsubsection{Information-Token Mask Construction}
\label{sec:information_masking}
We implement the masked language-model objective using \infstart{} and \infend{} as abbreviations for the \texttt{<information>} delimiters.

During batch collation, we compute $\mathbf{m}$ dynamically using token offset mappings provided by the tokenizer to locate character spans of \texttt{\infstart} ... \texttt{\infend} within each assistant response. Formally, for each conversation $c$ with assistant text $T_c$, we identify character-level spans
\[
\begin{gathered}
\mathcal{S}_c = \{(s_k, e_k)\}_{k=1}^{K_c}, \\
\text{where } T_c[s_k:e_k] \in [\infstart,\, \infend].
\end{gathered}
\]
Given the tokenizer offset map $\Omega_c = \{(s_i, e_i)\}_{i=1}^{L}$, a token $i$ is masked if and only if
\[
\begin{gathered}
\exists (s_k, e_k) \in \mathcal{S}_c \text{ such that } \\
(s_i < e_k) \wedge (e_i > s_k).
\end{gathered}
\]
The resulting per-conversation mask tensor $\mathbf{m}_c$ is concatenated across all conversations to form the final batch-level mask tensor $\mathbf{M} \in \{0,1\}^{B\times L}$ used to gate loss terms.

\subsection{Two-stage Training and Data Construction}
\label{sec:two_stage_training}

\begin{figure}[!htbp]
    \centering
\begin{minipage}{.90\textwidth}
\centering
    \includegraphics[width=\linewidth]{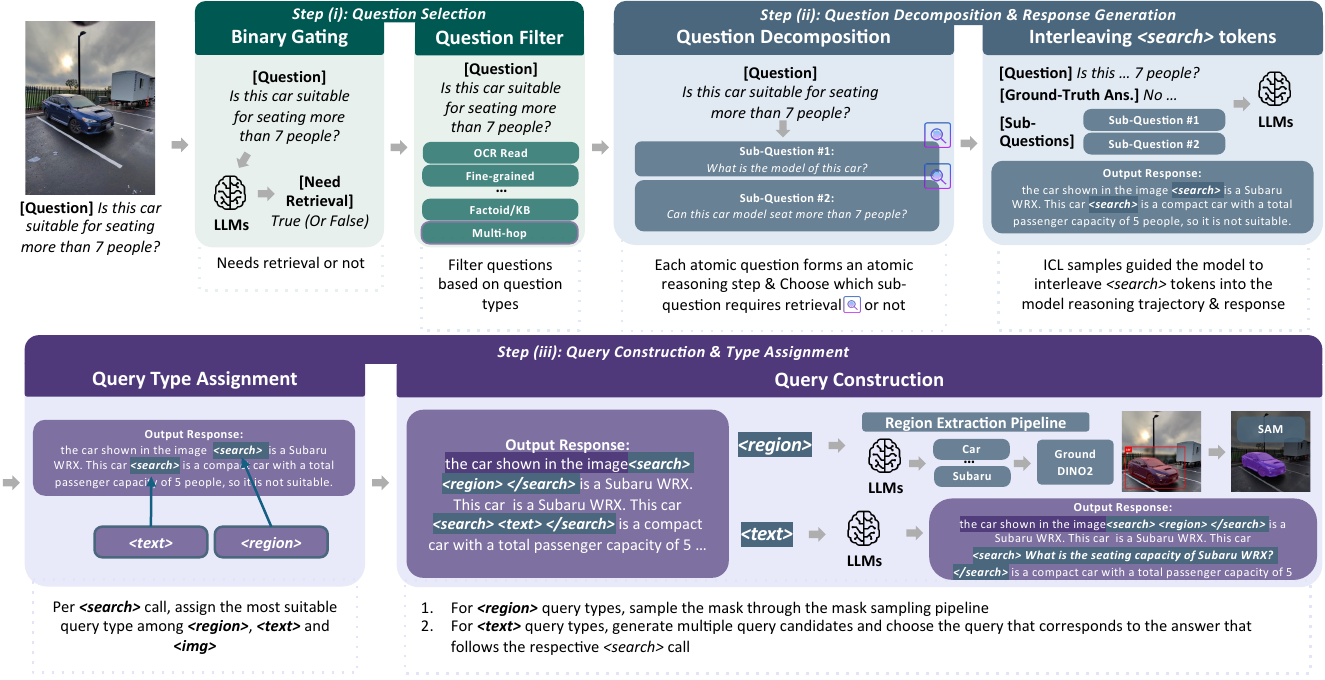}
    \caption{\textbf{Data construction pipeline for Stage-2 training.}}
    \label{fig:data_construction}
\end{minipage}
\end{figure}

\label{sec:training:implementation}
We initialize the LLaVA-13B backbone~\citep{liu2023llava} and text-aligned mask decoder from PLUM~\citep{blume-kim-2025partonomy}, then train in two stages to prevent segmentation forgetting while learning retrieval control.

\label{sec:training_recipe}

\paragraph{Stage 1: Mask Generation.} The mask segmentation performance can suffer from a catastrophic forgetting issue (\S \ref{sec:mask_segmentation}), requiring us to construct a training dataset that enables \textsc{PixSearch} to retain its visual reasoning and mask generation capabilities. We include the following datasets into the mixture for mask generation to enable segmentation on the nuanced language: ADE20k \citep{zhou2017scene}, Pascal Parts \citep{chen2014pascalpart}, PartImageNet \citep{he2022partimagenet}, PACO-LVIS \citep{ramanathan2023paco}, COCO-Stuff \citep{caesar2018cocostuff}, along with RefCOCO variants \citep{kazemzadeh-etal-2014-referitgame}. Our training mixture also employs a visual instruction tuning dataset from LLaVA \citep{liu2023llava}, which consists of 665k textual responses and captions given an image, enabling our segmenting LMM to retain its general visual understanding and reasoning ability.

\paragraph{Stage 2: Search-Interleaved Reasoning.} Training in Stage 2 aims to teach our model when to trigger search and when to construct a visually-grounded multimodal query (i.e., \texttt{<region>}, \texttt{<image>}). Consider the example question \textit{``What is the conservation status of this animal?''}, generating a textual caption of the long-tail animal directly for retrieval, instead of composing a multi-modal search query, could lead to hallucination \citep{kim-ji-2024-finer}.

Figure~\ref{fig:data_construction} depicts the process of training data generation for Stage 2. The documented training mixture uses TextVQA \citep{singh2019textvqa}, InfoSeek \citep{chen2023infoseek}, OVEN \citep{hu2023oven}, OKVQA \citep{marino2019okvqa}, and A-OKVQA \citep{schwenk2022aokvqa}. Additional WebQA \citep{chang2022webqa} records were generated but not included in this run. CRAG-MM \citep{wang2025crag} annotations and in-context examples were prepared separately and are not part of this mixture. For each sample, we construct training data in three steps.

(i) \textit{Question selection:} We prompt a proprietary LMM\footnote{we use \texttt{gpt-4.1} in this work} with in-context learning to determine if the question can benefit from external knowledge (e.g., \textit{Where is this plant native to?}) by classifying the question into 10 pre-defined VQA question types (Appendix~\ref{app:question_types}). Multi-hop External Knowledge Reasoning, Fine-grained Entity Identification, and Factoid/KB Questions are targeted for retrieval, which helps identify visual entities and obtain the external knowledge needed to answer the questions. The final mixture also includes examples without search calls.

(ii) \textit{Question decomposition \& response generation:} We decompose questions into multiple atomic sub-questions and determine independently per sub-question whether retrieval is needed.
Then, we feed the original question, sub-questions, and ground-truth answers into the prompt, and instruct LMMs to generate a \texttt{<search>}-interleaved reasoning trajectory (i.e., response). Here, we generate $N$ ($N=5$) such trajectories and feed it back to the model for self-refinement loop \citep{madaan2023self}, selecting the best response where the \texttt{<search>} token was appropriately placed in the reasoning trajectory when search is deemed necessary. 

(iii) \textit{Query construction \& type assignment:} Guided by the ICL samples, the model assigns a pseudo-gold query type to each \texttt{<search>} token that appears in the final response. We assign \texttt{<region>} when the question refers to an entity present in the input image, and assign \texttt{<image>} if the question requires a holistic understanding of the whole image. For text queries, we prompt LLMs to generate based on the immediately preceding text before the corresponding \texttt{<search>} token.

\subsubsection{Training Data Statistics}
\label{sec:training_statistics}
Table~\ref{tab:training_data_counts} reports counts from the saved generated training files and the dataset selection recorded for the completed Stage-2 run. The selected search-interleaved component contains 35,000 records. Records can repeat source questions; unique IDs are counted separately within each dataset, totaling 31,178 across the five selected datasets. These counts include examples both with and without search calls and do not represent the size of the full segmentation and visual-instruction mixture. The separate CRAG-MM annotation file contains 1,230 records; no corresponding final search-interleaved training file was found among the retained artifacts.

\begin{table}[!htbp]
\centering
\begin{minipage}{.63\textwidth}
\centering
\small
\setlength{\tabcolsep}{6pt}
\begin{tabular}{@{}lrrc@{}}
\toprule
Dataset & Records & Unique IDs & Used \\
\midrule
OVEN & 7,000 & 5,277 & \cmark \\
InfoSeek & 7,000 & 6,045 & \cmark \\
A-OKVQA & 7,000 & 6,841 & \cmark \\
OKVQA & 7,000 & 6,078 & \cmark \\
TextVQA & 7,000 & 6,937 & \cmark \\
\midrule
Selected total & 35,000 & 31,178 & \\
\midrule
WebQA & 7,000 & 6,064 & \xmark \\
\bottomrule
\end{tabular}
\caption{\textbf{Generated training-data statistics.} ``Used'' indicates inclusion in the documented Stage-2 run. Unique counts refer to source IDs within each dataset, not a deduplicated training mixture.}
\label{tab:training_data_counts}
\end{minipage}
\end{table}

\FloatBarrier

\section{Experiments}
\label{sec:experiment}
Through comprehensive experiments, we answer the following research questions.

\noindent
{\bf RQ1:} Is \textsc{PixSearch} effective on question answering across factual VQA, general VQA, and text-only QA?

\noindent
{\bf RQ2:} How much do region-level retrieval and multi-step search contribute to \textsc{PixSearch}'s performance?

\noindent
{\bf RQ3:} Does \textsc{PixSearch} preserve the segmentation capability? 

\subsection{Experiment setup}
{\bf Dataset:} We experimented with four VQA benchmarks CRAG-MM~\citep{wang2025crag}, TextVQA~\citep{singh2019textvqa}, InfoSeek~\citep{chen2023infoseek}, OVEN~\citep{hu2023oven}; text-only QA benchmarks are detailed in Appendix~\ref{app:text_qa}. We constructed our own search API using 6M Wikipedia documents\footnote{Derived from: Wikidump 2022/10/01} and their corresponding images, with DINOv3 \citep{simeoni2025dinov3} as our image encoder and MPNet\footnote{We use the \texttt{all-mpnet-base-v2}} as the text encoder. Each embedding in the image index is linked to a corresponding Wikipedia document such that an image-to-image retrieval returns the corresponding document. For CRAG-MM~\citep{wang2025crag}, we use the search API provided by the benchmark\footnote{\url{https://www.piwheels.org/project/cragmm-search-pipeline/}}. 

\paragraph{Evaluation:} Following previous work, we evaluated relaxed Exact Match (EM) and F1 score~\citep{kim-ji-2024-finer}. For CRAG-MM, we use the \textit{Truthfulness} score proposed by the benchmark \citep{wang2025crag}, where each answer is classified into \textit{accurate} (score=1), \textit{missing} (score=0) and \textit{hallucination} (score=-1), and truthfulness is the average score $\tau \in[-1.0, 1.0]$; the Truthfulness score is evaluated with LLM-as-Judge, with GPT-4.1 being the judge of the generated response. In Table \ref{tab:cragmm_full}, we multiply Truthfulness by 100 for consistency with rest of the other metrics on the table.

\paragraph{Implementations:} We compared \textsc{PixSearch} with LLM-only solutions (LLaVA-13B \citep{liu2023llava}, PLUM-13B~\citep{blume-kim-2025partonomy}, Llama-3.2-11B~\cite{grattafiori2024llama}), MM-RAG solutions using the whole image for search. We also separately implement GroundedSAM~\citep{ren2024grounded} + LLaVA-13B baseline so we can test against a stronger pipeline baseline; pipeline methods such as REVIVE~\citep{lin2022revive} rely on weaker grounding models such as GLIP~\citep{li2022grounded} and FiD~\citep{izacard2021leveraging}. For a state-of-the-art pipeline method, we also employ VLM-PRF \citep{hong2026knowledgebased} as another baseline. We detail the implementation of the pipeline in Appendix. We also compared with ablated version of our \textsc{PixSearch} model, including (i) No Search: the finetuned \textsc{PixSearch} model without search triggering; (ii) Text Question: always using the text question as the search query, wherein the retrieval results are prepended to the input question for output generation; (iii) Whole Image: using the input image as the search query instead.

\begin{table}[!htbp]
\small
\centering
\setlength{\tabcolsep}{1.5pt}
\renewcommand{\arraystretch}{1.15}

\begin{adjustbox}{max width=\textwidth}
\begin{tabular}{l|cccc|cccc|cccc}
\toprule
\multirow{2}{*}{\textbf{Models}} &
\multicolumn{4}{c|}{\textbf{CRAG-MM (Overall)}} &
\multicolumn{4}{c|}{\textbf{CRAG-MM (Egocentric)}} &
\multicolumn{4}{c}{\textbf{CRAG-MM (Non-Egocentric)}} \\
\cmidrule(lr){2-5}\cmidrule(lr){6-9}\cmidrule(lr){10-13}
& \textbf{Truth. $\uparrow$} & \textbf{Acc. $\uparrow$} & \textbf{Miss. $\downarrow$} & \textbf{Hallu. $\downarrow$}
& \textbf{Truth. $\uparrow$} & \textbf{Acc. $\uparrow$} & \textbf{Miss. $\downarrow$} & \textbf{Hallu. $\downarrow$}
& \textbf{Truth. $\uparrow$} & \textbf{Acc. $\uparrow$} & \textbf{Miss. $\downarrow$} & \textbf{Hallu. $\downarrow$} \\
\midrule

\rowcolor{groupgray}\multicolumn{13}{c}{\textit{Llama-3.2-11B-Vision}~\citep{grattafiori2024llama}} \\

Llama-3.2-11B$_{\texttt{No Search}}$ 
& -16.9 & 24.4 & 34.4 & 41.3
& -22.5 & 21.0 & 35.5 & 43.5
& 0.1 & 34.2 & 31.7 & \textbf{34.1} \\

Llama-3.2-11B$_{\texttt{Whole Image}}$ 
& -8.6 & 35.3 & 20.3 & 43.9
& -17.0 & 31.0 & 21.0 & 48.0
& \textbf{11.5} & \textbf{46.6} & 18.3 & 35.1 \\
\midrule

\rowcolor{groupgray}\multicolumn{13}{c}{\textit{Additional MM-RAG Baselines}} \\

\makecell[l]{GroundedSAM+LLaVA-13B\\\citep{ren2024grounded}} 
& -30.4 & 23.4 & 22.8 & 53.8 
& -41.7 & 18.1 & 22.0 & 59.8 
& -23.6 & 25.7 & 25.0 & 49.3 \\

\makecell[l]{VLM-PRF (Qwen-2.5-7B)\\\citep{hong2026knowledgebased}} 
& -23.6 & 25.8 & 24.7 & 49.5
& -31.5 & 21.6 & 25.4 & 53.0
& -2.8 & 37.3 & 22.7 & 40.1 \\
\makecell[l]{ReflectiVA-8B\\\citep{cocchi2025reflectiva}}
& -47.0 & 24.1 & 4.7 & 71.1
& -50.0 & 22.6 & 4.7 & 72.7
& -38.4 & 28.3 & 4.9 & 66.8 \\

\makecell[l]{EchoSight (Llama-3-8B)\\\citep{yan2024echosight}}
& -71.4 & 12.9 & 2.8 & 84.3
& -76.7 & 10.0 & 3.2 & 86.8
& -56.5 & 20.8 & 1.9 & 77.3 \\
\midrule

\rowcolor{groupgray}\multicolumn{13}{c}{\textit{End-to-End MM-RAG}} \\

LLaVA-13B~\citep{liu2023llava} 
& -69.0 & 10.3 & \textbf{10.4} & 79.3
& -72.8 & 8.1 & \textbf{11.0} & 80.9
& -59.1 & 16.1 & \textbf{8.7} & 75.2 \\

\makecell[l]{PLUM-13B\\\citep{blume-kim-2025partonomy}} 
& -35.2 & 25.3 & 14.5 & 60.5
& -43.2 & 20.5 & 15.8 & 63.7
& -10.1 & 39.4 & 11.0 & 49.5 \\

PixSearch-13B$_{\texttt{No Search}}$ 
& $-40.6$ & $19.6$ & $20.0$ & $60.2$
& $-48.0$ & $15.5$ & $20.8$ & $63.5$
& $-19.7$ & $30.8$ & $17.8$ & $50.5$ \\

PixSearch-13B$_{\texttt{Text Query}}$ 
& $-44.5$ & $18.2$ & $19.1$ & $62.7$
& $-52.5$ & $12.2$ & $20.1$ & $67.7$
& $-14.8$ & $34.4$ & $16.4$ & $49.2$ \\

PixSearch-13B$_{\texttt{Whole Image}}$ 
& $-19.3$ & $31.2$ & $18.1$ & $50.5$
& $-27.8$ & $26.5$ & $19.0$ & $54.3$
& $4.5$ & $44.2$ & $15.5$ & $39.7$ \\

PixSearch-13B$_{\texttt{Full}}$ 
& $\textbf{-3.8}$ & $\textbf{36.9}$ & $21.6$ & $\textbf{40.7}$
& $\textbf{-12.3}$ & $\textbf{32.4}$ & $22.3$ & $\textbf{44.7}$
& $10.2$ & $45.6$ & $18.7$ & $35.4$ \\

\bottomrule
\end{tabular}
\end{adjustbox}
\caption{CRAG-MM results: all metrics in \%, with Truthfulness $\in [-100,100]$. GroundedSAM+LLaVA-13B uses \texttt{gpt-4.1} for entity extraction.}
\label{tab:cragmm_full}
\end{table}

\subsection{Overall performance (RQ1, RQ2)}
\label{sec:search_interleaved_reasoning}

Table~\ref{tab:cragmm_full} compares \textsc{PixSearch}-13B$_{\text{Full}}$ against VQA and MM-RAG baselines and ablations. PixSearch obtains 36.9\% overall accuracy and $-3.8$ truthfulness. Across all three splits, PixSearch yields more missing answers than LLaVA-13B but substantially fewer hallucinations, indicating a trade-off between answer coverage and factual reliability. One plausible explanation is that retrieval-conditioned reasoning encourages abstention when evidence is insufficient, whereas LLaVA's lower missing-answer rate partly reflects unsupported answers rather than successful responses. Second, \textsc{PixSearch}$_{\text{Full}}$ surpasses Llama-3.2 in overall accuracy and truthfulness for both LLM-only and whole-image MM-RAG settings. Third, \textsc{PixSearch}$_{\text{Text Query}}$ performs similarly or worse than \textsc{PixSearch}$_{\text{No Search}}$, confirming that text-only retrieval is ineffective, especially for egocentric images. Fourth, although \textsc{PixSearch}$_{\text{Whole Image}}$ improves over no-search, the full model, which is capable of multi-step and region-level retrieval, achieves substantially better performance. Compared to the pipelined GroundedSAM+LLaVA approach, \textsc{PixSearch}$_{\text{Full}}$ performs significantly better, highlighting the benefits of interleaving text and mask-based reasoning. Finally, we observe a large performance gap between egocentric and non-egocentric images, underscoring the difficulty of egocentric VQA with smaller models. Section~\ref{sec:qualitative_case_studies} provides qualitative examples.

\subsection{Mask Segmentation (RQ3)}
\label{sec:mask_segmentation}

\begin{table}[!htbp]
\centering
\begin{minipage}{.78\textwidth}
\centering
\small
\setlength{\tabcolsep}{6pt}
\textbf{(a) Referring expression segmentation.}\par\smallskip
\begin{adjustbox}{max width=\columnwidth}
\begin{tabular}{@{}lcccc@{}}
\toprule
& LISA~7B & GLaMM & \multicolumn{2}{c}{\textsc{PixSearch}} \\
\cmidrule(lr){4-5}
Dataset / split & & & 1:9 & 7:9 \\
Extra Seg. & \xmark & \cmark & \xmark & \xmark \\
\midrule
RefCOCO val & 74.1 & \textbf{79.5} & 15.3 & \textit{78.9} \\
RefCOCO testA & 76.5 & \textbf{83.2} & 13.4 & \textit{79.3} \\
RefCOCO testB & 71.1 & \textbf{76.9} & 11.2 & \textit{72.7} \\
RefCOCOg val(U) & 66.4 & \textbf{74.2} & 9.7 & \textit{69.8} \\
RefCOCOg test(U) & 68.5 & \textbf{74.9} & 10.2 & \textit{70.9} \\
RefCOCO+ val & 62.4 & \textbf{72.6} & 13.6 & \textit{65.9} \\
RefCOCO+ testA & 67.4 & \textbf{78.7} & 12.3 & \textit{69.9} \\
\bottomrule
\end{tabular}
\end{adjustbox}
\par\medskip
\textbf{(b) Semantic segmentation.}\par\smallskip
\begin{tabular}{@{}lcc@{}}
\toprule
Model & ADE20K & COCOStuff \\
\midrule
MaskFormer & 52.70 & 47.20 \\
PLUM & 55.08 & \textbf{49.97} \\
\textsc{PixSearch} (1:9) & 9.85 (-82.1\%) & 11.07 (-77.8\%) \\
\textsc{PixSearch} (5:9) & 47.19 (-14.3\%) & 42.55 (-14.9\%) \\
\textsc{PixSearch} (7:9) & \textbf{55.98} (+1.6\%) & 49.81 (-0.3\%) \\
\bottomrule
\end{tabular}
\caption{\textbf{Segmentation experiments.}
(a) Referring expression segmentation performance in gIoU. 
(b) Semantic segmentation performance in mIoU. 
\textsc{PixSearch} preserves competitive segmentation ability, with the best performance obtained at the 7:9 Stage-1/Stage-2 sampling ratio. Baselines: LISA~\citep{lai2024lisa}, GLaMM~\citep{rasheed2024glamm}, MaskFormer~\citep{cheng2021maskformer}, and PLUM~\citep{blume-kim-2025partonomy}.}
\label{tab:segmentation_results}
\end{minipage}
\end{table}

For semantic segmentation, we evaluate on ADE20K~\citep{zhou2019semantic} and COCOStuff~\citep{caesar2018cocostuff}. Table~\ref{tab:segmentation_results}(b) shows that \textsc{PixSearch} achieves comparable performance to MaskFormer. In referring expression segmentation (Table~\ref{tab:segmentation_results}(a)), \textsc{PixSearch} also performs competitively with strong segmenting LMMs such as GLaMM~\citep{rasheed2024glamm}, and outperforms variants like LISA. We observe that the Stage-1/Stage-2 sampling ratio significantly affects mask prediction performance; in our experiments, a 7:9 ratio yields the best results. These findings suggest that \textsc{PixSearch}'s mask decoder can function as a standalone module for pixel-level grounding of noun phrases and referring expressions.

\begin{table}[!htbp]
\centering
\begin{minipage}{.84\textwidth}
\centering
\small
\setlength{\tabcolsep}{6pt}
\renewcommand{\arraystretch}{1.1}
\begin{tabular}{@{}lccc@{}}
\toprule
\textbf{Model (13B)} & \makecell{\textbf{TextVQA}\\EM / F1} & \makecell{\textbf{InfoSeek}\\EM / F1} & \makecell{\textbf{OVEN}\\EM / F1} \\
\midrule
LLaVA & 22.84\,/\,26.15 & 13.94\,/\,14.98 & 2.20\,/\,3.12 \\
PLUM & 30.11\,/\,32.62 & 15.60\,/\,16.54 & 2.64\,/\,3.81 \\
PixSearch$_{\texttt{No Search}}$ & 33.0\,/\,34.9 & 18.6\,/\,19.7 & 4.0\,/\,5.6 \\
PixSearch$_{\texttt{Question}}$ & 31.2\,/\,33.3 & 21.0\,/\,22.6 & 5.5\,/\,7.0 \\
PixSearch$_{\texttt{Whole Image}}$ & 32.6\,/\,34.5 & 24.6\,/\,26.4 & 15.6\,/\,17.3 \\
PixSearch$_{\texttt{Full}}$ & 32.8\,/\,\textbf{35.5} & \textbf{24.9}\,/\,\textbf{27.0} & \textbf{16.2}\,/\,\textbf{17.4} \\
\bottomrule
\end{tabular}
\caption{\textbf{Performance on additional VQA benchmarks.}
Scores are reported in \%. For PixSearch variants, we report mean scores over three independent training seeds. Each cell shows EM / F1; all models have 13B parameters.}
\label{tab:vqa_other}
\end{minipage}
\end{table}

\paragraph{Generalization across VQA and text QA.}
Across diverse VQA benchmarks (Table~\ref{tab:vqa_other}), \textsc{PixSearch} slightly outperforms PLUM-13B on TextVQA, where retrieval is unnecessary, and consistently improves over its non-RAG backbone on knowledge-intensive datasets such as InfoSeek and OVEN. For these entity-centric images, whole-image retrieval (PixSearch$_{\text{Whole Image}}$) is sufficient and does not degrade performance. Text-only QA results and analysis are provided in Appendix~\ref{app:text_qa}.

\subsection{Search Pattern Analysis (RQ2)}

\begin{figure}[!htbp]
\centering
\begin{minipage}[t]{.47\textwidth}
\vspace{0pt}
\centering
\includegraphics[width=.82\linewidth]{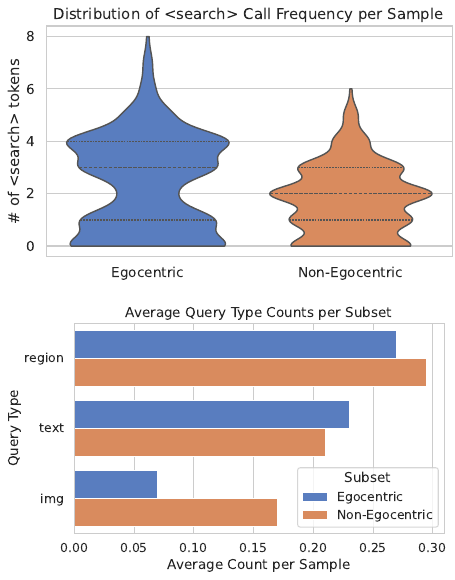}
\caption{\textbf{Search behavior on CRAG-MM.} Search behavior of \textsc{PixSearch}$_{\text{Interleaved}}$, showing search-call frequency (top) and mean query-type counts (bottom).}
\label{fig:search_behavior}
\label{fig:search_and_retrieval}
\end{minipage}\hfill
\begin{minipage}[t]{.47\textwidth}
\vspace{0pt}
\centering
\includegraphics[width=\linewidth]{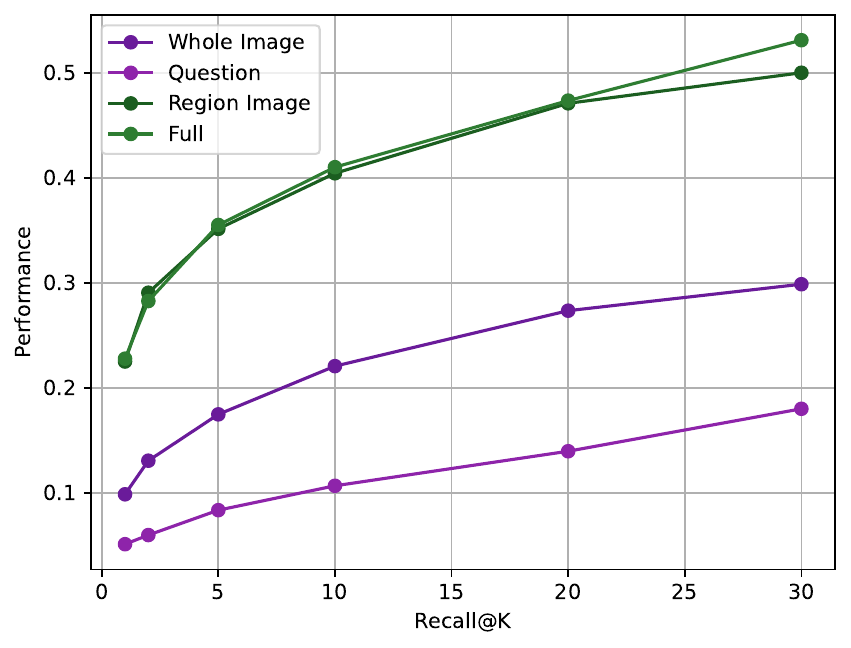}
\caption{\textbf{Retrieval performance (Recall@K) on CRAG-MM.} Region-level queries outperform whole-image and text-based queries, especially at low $K$.}
\label{fig:retrieval_performance}
\end{minipage}
\end{figure}

In CRAG-MM, we split images into egocentric and non-egocentric subsets and analyze (i) search frequency via the number of \searchtag{} tokens and (ii) the distribution of \regiontag, \imagetag, and text queries. As shown in Figure~\ref{fig:search_behavior}, egocentric images trigger more searches than non-egocentric ones (3--4 vs. 2--3), suggesting greater need for region-specific evidence and reasoning. Both subsets frequently use \regiontag, while text queries account for about 24\% of egocentric searches, often as follow-ups. Using ground-truth entities and aliases, we further show that region-level queries outperform whole-image and text retrieval, especially at Recall@1 and Recall@5 (Figure~\ref{fig:retrieval_performance}). Section~\ref{sec:ablations} studies query modalities and search budgets; Section~\ref{sec:formal_decode} formalizes decoding.

\begin{table}[!htbp]
\centering
\begin{minipage}{.82\textwidth}
\centering
\small
\setlength{\tabcolsep}{6pt}
\begin{tabular}{@{}lcc@{}}
\toprule
\textbf{Method} & \textbf{Time (s)} & \textbf{p90 (s)} \\
\midrule
\textsc{PixSearch-13B} (Full) & $2.8\pm0.9$ & 3.3 \\
\textsc{PixSearch-13B} (Whole Image) & $\mathbf{2.45}\pm0.72$ & $\mathbf{2.96}$ \\
GroundedSAM+LLaVA-13B$^{\dagger}$ & $4.4\pm1.3$ & 4.1 \\
GroundedSAM+LLaVA-13B$^{\ddagger}$ & $5.0\pm3.4$ & 5.9 \\
\bottomrule
\end{tabular}
\caption{\textbf{Inference latency.} Wall-clock latency on CRAG-MM, reported as mean $\pm$ std over five runs. $\dagger$: cached, GPU-only pipeline. $\ddagger$: pipeline including GPT-4.1 entity extraction.}
\label{tab:latency}
\end{minipage}
\end{table}

\begin{table}[!htbp]
\centering
\begin{minipage}{.64\textwidth}
\centering
\small
\setlength{\tabcolsep}{6pt}
\begin{tabular}{@{}lcc@{}}
\toprule
\textbf{Training} & \textbf{Acc.} & \textbf{gIoU} \\
\midrule
Stage-2 SFT only & 18.5 & 14.8 \\
Single-stage joint SFT & 32.6 & 67.4 \\
Two-stage SFT & \textbf{36.9} & \textbf{78.9} \\
Search-R1-style GRPO & \textbf{36.9} & 71.6 \\
\bottomrule
\end{tabular}
\caption{\textbf{Training comparison.} CRAG-MM accuracy and RefCOCO gIoU (\%). Expanded metrics are in Appendix~\ref{app:training_controls}.}
\label{tab:stage1_ablation}
\end{minipage}
\end{table}

\paragraph{Training comparison.}
Table~\ref{tab:stage1_ablation} compares two-stage SFT with Stage-2-only SFT, joint SFT, and a Search-R1-style GRPO control~\citep{jin2025searchr1}. Two-stage SFT improves accuracy over joint SFT by 4.3 points. It ties GRPO at 36.9\% accuracy while retaining higher RefCOCO gIoU (78.9 vs.\ 71.6), supporting a grounding-retention advantage in this setting rather than an accuracy advantage over RL.

\paragraph{Inference latency.}
Table~\ref{tab:latency} shows that \textsc{PixSearch} reduces mean latency by 36.4\% relative to the cached GroundedSAM--LLaVA pipeline (2.8 vs.\ 4.4\,s), with lower reported variability and p90 latency. Relative to the pipeline including GPT-4.1 entity extraction, the reduction is 44\% (2.8 vs.\ 5.0\,s). Whole-image retrieval is 0.35\,s faster than Full, showing the cost of region-grounded search. Appendix~\ref{app:latency_breakdown} gives component timings and retrieval-call counts.

\subsection{Ablation Studies (RQ2)}
\label{sec:ablations}
\paragraph{Query modality removal.} 
Table~\ref{tab:ablation_question_type} evaluates the contribution of different search query modalities by restricting the types of queries \textsc{PixSearch} can issue at inference time. Among single-modality settings, \textsc{PixSearch}$_{\texttt{Only Region}}$ performs closest to the full model, especially on egocentric images, confirming that pixel-grounded region retrieval is the primary driver of performance in visually cluttered, entity-centric scenes. In contrast, \textsc{PixSearch}$_{\texttt{Only Text}}$ shows the largest degradation, with substantially lower accuracy and higher hallucination, indicating that text-only queries are often insufficient for precise entity retrieval.

Whole-image retrieval (\textsc{PixSearch}$_{\texttt{Only Image}}$) performs better than text-only but remains clearly inferior to region-based search, highlighting the limitations of coarse visual queries. Removing text queries (\textsc{PixSearch}$_{\texttt{No Text}}$) results in only a small overall accuracy drop relative to the full model, suggesting that textual queries mainly serve as complementary follow-up searches. Similarly, removing whole-image queries (\textsc{PixSearch}$_{\texttt{No Image}}$) causes a modest degradation, particularly on non-egocentric images, where global context can be helpful.

\begin{table}[!htbp]
\small
\centering
\setlength{\tabcolsep}{1.2pt}
\renewcommand{\arraystretch}{1.15}

\begin{adjustbox}{max width=\linewidth}
\begin{tabular}{l|cccc|cccc|cccc}
\toprule
\multirow{2}{*}{\textbf{Models}} &
\multicolumn{4}{c|}{\textbf{CRAG-MM (Overall)}} &
\multicolumn{4}{c|}{\textbf{CRAG-MM (Egocentric)}} &
\multicolumn{4}{c}{\textbf{CRAG-MM (Non-Egocentric)}} \\
\cmidrule(lr){2-5}\cmidrule(lr){6-9}\cmidrule(lr){10-13}
& \textbf{Truth. $\uparrow$} & \textbf{Acc. $\uparrow$} & \textbf{Miss. $\downarrow$} & \textbf{Hallu. $\downarrow$}
& \textbf{Truth. $\uparrow$} & \textbf{Acc. $\uparrow$} & \textbf{Miss. $\downarrow$} & \textbf{Hallu. $\downarrow$}
& \textbf{Truth. $\uparrow$} & \textbf{Acc. $\uparrow$} & \textbf{Miss. $\downarrow$} & \textbf{Hallu. $\downarrow$} \\
\midrule

\rowcolor{groupgray}\multicolumn{13}{c}{\textit{Question Type Removal (Inference-Time Constraint)}} \\

PixSearch-13B$_{\texttt{Full}}$ & \textbf{-3.8} & \textbf{36.9} & 21.6 & \textbf{40.7}
      & \textbf{-12.3} & \textbf{32.4} & 22.3 & \textbf{44.7}
      & \textbf{10.2} & \textbf{45.6} & 18.7 & \textbf{35.4} \\

PixSearch-13B$_{\texttt{No Region}}$
  & -14.5 & 31.9 & 21.7 & 46.4
  & -23.0 & 27.5 & 22.0 & 50.5
  & 8.0 & 43.5 & 21.0 & 35.5 \\

PixSearch-13B$_{\texttt{No Text}}$
  & -6.3 & 35.9 & 21.9 & 42.2
  & -11.9 & 32.8 & 22.5 & 44.7
  & 8.5 & 44.0 & 20.5 & 35.5 \\

PixSearch-13B$_{\texttt{No Image}}$
  & -7.5 & 34.7 & 23.1 & 42.3
  & -13.0 & 32.0 & 23.0 & 45.0
  & 7.0 & 42.0 & 23.0 & 35.0 \\

PixSearch-13B$_{\texttt{Only Region}}$
  & -6.4 & 35.9 & 21.9 & 42.3
  & -12.2 & 32.6 & 22.6 & 44.8
  & 9.0 & 44.5 & 20.0 & 35.5 \\

PixSearch-13B$_{\texttt{Only Text}}$
  & -44.4 & 18.3 & 19.1 & 62.6
  & -55.5 & 12.2 & 20.1 & 67.7
  & -14.8 & 34.3 & 16.5 & 49.2 \\

PixSearch-13B$_{\texttt{Only Image}}$
  & -18.7 & 31.7 & 18.0 & 50.3
  & -27.6 & 26.7 & 19.0 & 54.3
  & 5.1 & 44.8 & 15.5 & 39.7 \\
\bottomrule
\end{tabular}
\end{adjustbox}
\caption{
\textbf{Ablation study on question type removal.}
We restrict \textsc{PixSearch} from issuing specific types of search queries during decoding and measure the effect on performance. The full-model reference row uses Table~\ref{tab:cragmm_full} values and is highlighted in bold. All metrics are reported in \%.
}
\label{tab:ablation_question_type}
\end{table}

\paragraph{Search budget.}Table~\ref{tab:ablation_search_budget} studies how limiting the number of allowed search calls affects performance. Allowing even a single search substantially improves over the no-search baseline, demonstrating the importance of external knowledge retrieval. Performance continues to improve as the search budget increases, with most gains realized within two to three search calls, particularly for egocentric images that require iterative entity identification and reasoning.

Three to four searches approach the full model in overall accuracy, although subset accuracy and truthfulness are not monotonic in the search budget. The full model attains the highest overall accuracy, while some constrained settings yield slightly higher truthfulness or lower hallucination. These results indicate a trade-off between retrieval budget and answer quality.

\begin{table}[!htbp]
\small
\centering
\setlength{\tabcolsep}{1.2pt}
\renewcommand{\arraystretch}{1.15}

\begin{adjustbox}{max width=\linewidth}
\begin{tabular}{l|cccc|cccc|cccc}
\toprule
\multirow{2}{*}{\textbf{Models}} &
\multicolumn{4}{c|}{\textbf{CRAG-MM (Overall)}} &
\multicolumn{4}{c|}{\textbf{CRAG-MM (Egocentric)}} &
\multicolumn{4}{c}{\textbf{CRAG-MM (Non-Egocentric)}} \\
\cmidrule(lr){2-5}\cmidrule(lr){6-9}\cmidrule(lr){10-13}
& \textbf{Truth. $\uparrow$} & \textbf{Acc. $\uparrow$} & \textbf{Miss. $\downarrow$} & \textbf{Hallu. $\downarrow$}
& \textbf{Truth. $\uparrow$} & \textbf{Acc. $\uparrow$} & \textbf{Miss. $\downarrow$} & \textbf{Hallu. $\downarrow$}
& \textbf{Truth. $\uparrow$} & \textbf{Acc. $\uparrow$} & \textbf{Miss. $\downarrow$} & \textbf{Hallu. $\downarrow$} \\
\midrule

\rowcolor{groupgray}\multicolumn{13}{c}{\textit{Search Budget Constraint}} \\

  PixSearch-13B$_{\texttt{No Search}}$ ($B{=}0$) 
  & -40.4 & 19.8 & 20.0 & 60.2
          & -47.8 & 15.7 & 20.8 & 63.5
          & -18.8 & 31.7 & 17.8 & 50.5 \\

PixSearch-13B$_{\texttt{Search}\leq1}$
  & -7.9 & 32.1 & 21.8 & 43.3
  & -16.5 & 28.0 & 22.9 & 49.1
  & 15.0 & 43.0 & 18.9 & 28.0 \\

PixSearch-13B$_{\texttt{Search}\leq2}$
  & -3.5 & 35.8 & 21.6 & 39.9
  & -12.4 & 32.0 & 22.8 & 45.2
  & 20.0 & 46.0 & 18.4 & 26.0 \\

PixSearch-13B$_{\texttt{Search}\leq3}$
  & -3.0 & 36.4 & 21.5 & 39.5
  & -12.0 & 32.7 & 22.6 & 44.7
  & 20.8 & 46.2 & 18.5 & 25.7 \\

PixSearch-13B$_{\texttt{Search}\leq4}$
  & -2.9 & 36.5 & 21.4 & 39.4
  & -11.8 & 32.9 & 22.5 & 44.6
  & 20.6 & 46.0 & 18.6 & 25.6 \\

PixSearch-13B$_{\texttt{Full}}$ (unbounded) & \textbf{-3.8} & \textbf{36.9} & 21.6 & \textbf{40.7}
      & \textbf{-12.3} & \textbf{32.4} & 22.3 & \textbf{44.7}
      & \textbf{10.2} & \textbf{45.6} & 18.7 & \textbf{35.4} \\

\bottomrule
\end{tabular}
\end{adjustbox}
\caption{
\textbf{Ablation study on the number of search tokens.}
We limit the maximum number of \searchtag calls allowed during decoding and evaluate how multi-step search contributes to performance. The full-model reference row uses Table~\ref{tab:cragmm_full} values and is highlighted in bold. All metrics are reported in \%.
}
\label{tab:ablation_search_budget}
\end{table}

\subsection{Qualitative Case Studies}
\label{sec:qualitative_case_studies}

Table~\ref{tab:qualitative_examples} provides qualitative examples comparing \textsc{PixSearch} with baseline LMMs on CRAG-MM. These examples illustrate how \textsc{PixSearch} performs search-interleaved reasoning by first grounding the relevant entity with a predicted mask, retrieving region-level external evidence, and then issuing follow-up text queries when additional factual context is needed. Compared with baseline models that rely on parametric knowledge alone, \textsc{PixSearch} produces answers that are better grounded in the visual entity and retrieved evidence.

\begin{table}[!htbp]
\small
\centering
\setlength{\tabcolsep}{1pt}
\renewcommand{\arraystretch}{1.0}

\begin{adjustbox}{max width=\linewidth}
\begin{tabular}{@{}M{0.16\textwidth}|p{0.64\textwidth}|M{0.16\textwidth}@{}}
\toprule
\textbf{Input Image} &
\centering\textbf{Question \& Model Outputs} &
\textbf{\textsc{PixSearch} Mask Overlay} \\
\midrule

\includegraphics[width=0.8\linewidth]{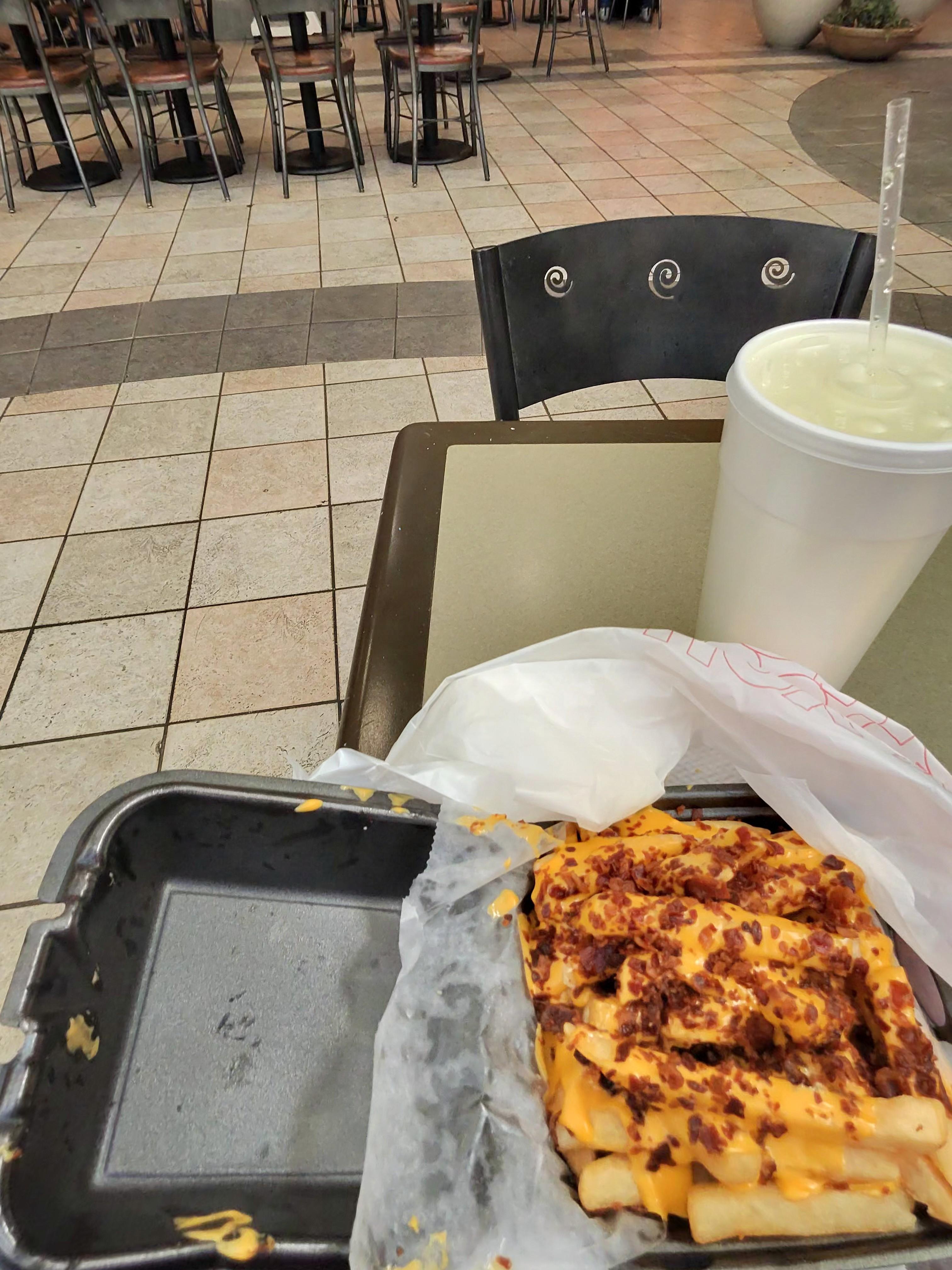} &
\makecell[{{p{\linewidth}}}]{
\textbf{Q:} What is the origin of this food item?\\[1pt]
\textbf{PLUM-13B:} It looks like a pasta with toppings and sauce on top of it.\\[3pt]
\textbf{\textsc{PixSearch}-13B$_{\text{Full}}$:} The origin of the food item shown in the image {\textcolor{Gray}{\em \searchtag \regiontag \searchend \informationtag Carne asada fries \textbar\ Bacon Cheeese Fries \textbar\ Restaurant \textbar\ ... \informationend}} is bacon cheese fries. The food item in the image {\textcolor{Gray}{\em\searchtag what is the origin of cheese bread or cheese fries?\ \searchend \informationtag Southwest US cheese fries \textbar\ McDonald's bacon cheese fries 2004 \textbar\ ... \informationend}} is a type of dish that originated in the United States.\\[1pt]
\textbf{GT:} Bacon cheese fries are from the United States, but the exact origin of the dish is not known.
} &
\includegraphics[width=0.8\linewidth]{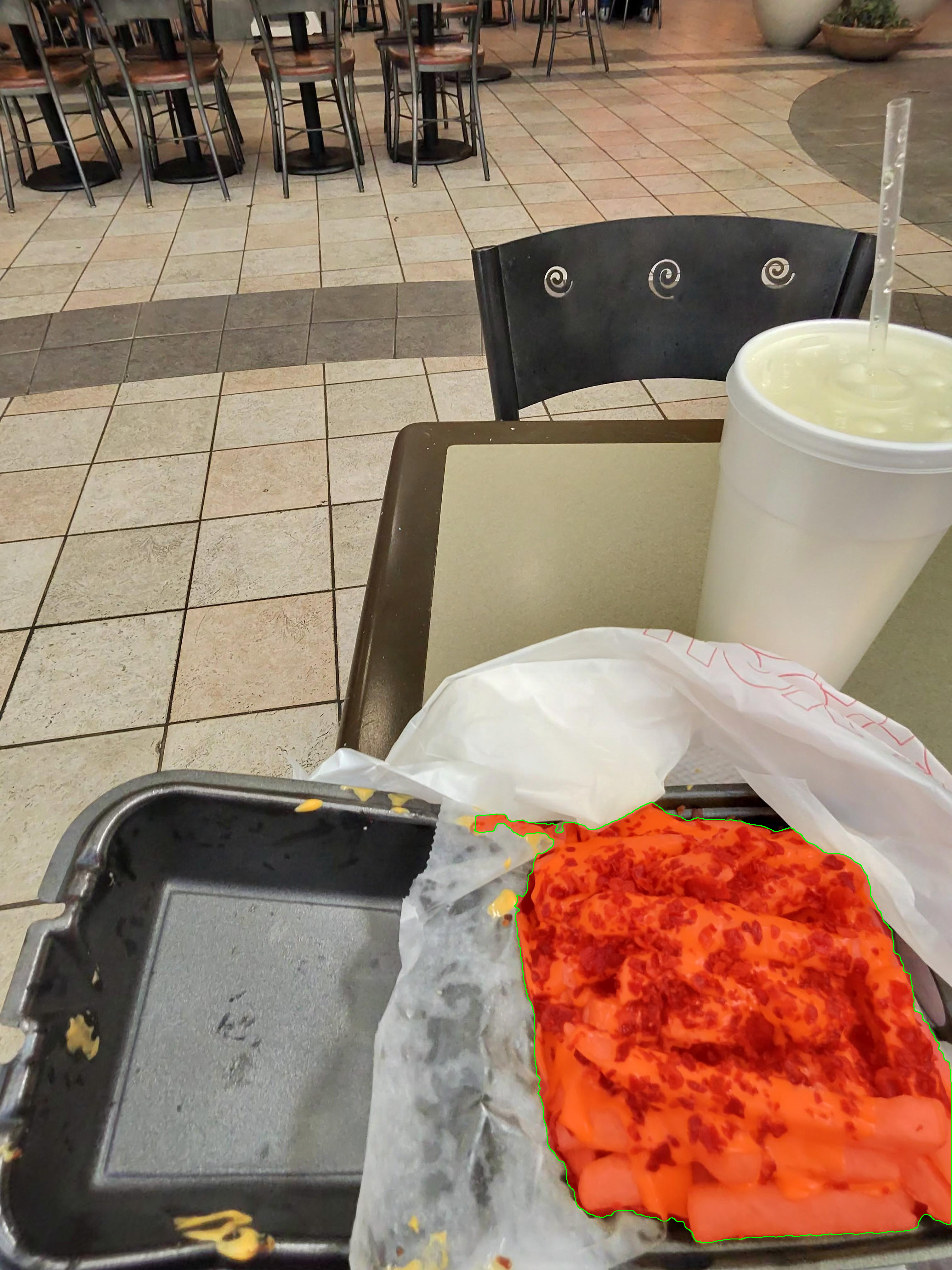} \\
\midrule

\includegraphics[width=0.8\linewidth]{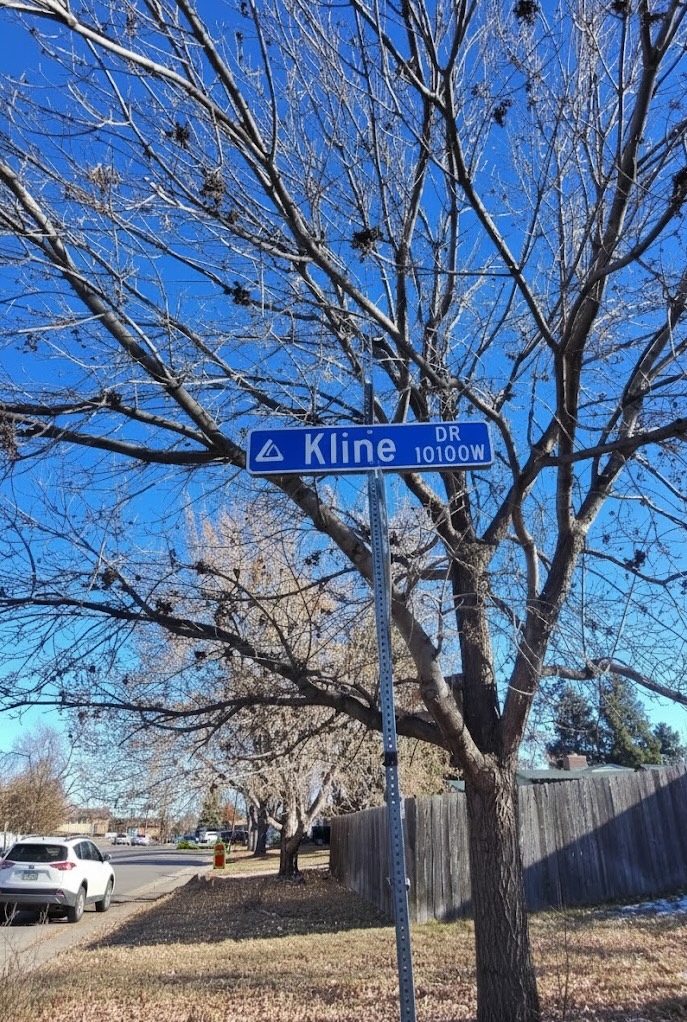} &
\makecell[{{p{\linewidth}}}]{
\textbf{Q:} Did this vehicle appear in the movie Pretty Woman back in the 90s?\\[1pt]
\textbf{LLaVA-13B:} The white hatchback did not appear in the movie Pretty Woman.\\[1pt]
\textbf{\textsc{PixSearch}-13B$_{\text{Full}}$} The vehicle shown in the image is a white SUV. This vehicle {\textcolor{Gray}{\em\searchtag \regiontag \searchend \informationtag Lincoln MKS 2009-2016 \textbar\ V6 engine 3.7L 3.5L \textbar\ Toyota Rav4 2014 \textbar\ Toyota RAV4 \textbar\ Chicago Assembly \textbar\ all-wheel drive \informationend}} is a Toyota RAV4. The Pretty Woman was released {\textcolor{Gray}{\em\searchtag when was the Pretty Woman released? \searchend \informationtag Pretty Woman \textbar\ Notting Hill \textbar\ Movies in 90s \informationend}} in 1989. The Toyota RAV4 was launched {\textcolor{Gray}{\em\searchtag when did the Toyota RAV4 launch? \searchend \informationtag Toyota RAV4 \textbar\ Toyota \textbar\ Toyota Motor Corporation \textbar\ Japan Car of the Year \textbar\ gasoline engine \informationend}} in 1994. Therefore, this vehicle is not the vehicle that appears in the movie Pretty Woman.\\[1pt]
\textbf{GT:} No, because the first-generation Toyota RAV 4 was launched in 1994, four years after the release date of the movie Pretty Woman on March 23, 1990.
} &
\includegraphics[width=0.9\linewidth]{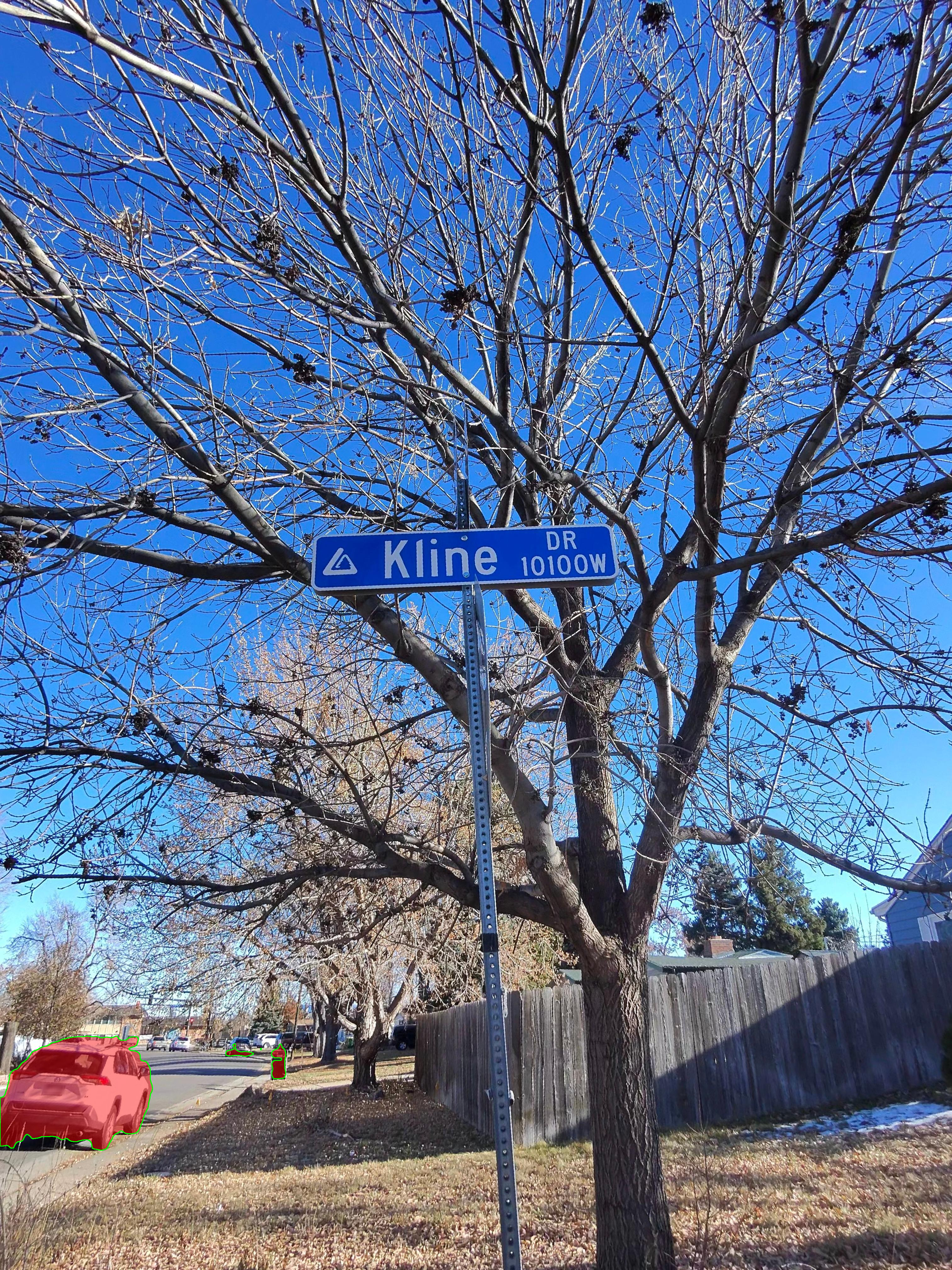} \\
\bottomrule
\end{tabular}
\end{adjustbox}
\caption{\textbf{Qualitative comparison between baseline and \textsc{PixSearch}-Interleaved outputs.}
Each example shows the input image (left), the corresponding question and model responses (middle), and the pixel-level grounding mask generated by the model (right). The payloads between the \informationtag \informationend are truncated to show entity names / titles only for brevity.}
\label{tab:qualitative_examples}
\end{table}

\section{Conclusion \& Future Works}
\label{sec:conclusion}

We present \textsc{PixSearch}, an end-to-end segmenting Large Multimodal Model for region-level retrieval-augmented reasoning. Motivated by egocentric VQA in wearable settings, where questions often require knowledge about a specific small or long-tail entity rather than the full scene, \textsc{PixSearch} learns when retrieval is needed, how to route queries across text, whole-image, and region-level search, and how to ground answers in retrieved evidence. Unlike modular pipelines that rely on external detectors, segmenters, or captioners, \textsc{PixSearch} integrates mask generation and retrieval control into a single autoregressive reasoning process.

Experiments show that this design is especially effective when retrieval must be aligned with fine-grained visual evidence. On CRAG-MM, \textsc{PixSearch} improves accuracy and reduces hallucination over whole-image retrieval and pipeline baselines, with particularly strong gains on egocentric images. Ablations show that region-level retrieval and multi-step search drive these improvements: text-only retrieval is often insufficient, whole-image retrieval helps but remains noisy, and region-based search provides the strongest grounding signal. At the same time, \textsc{PixSearch} preserves competitive segmentation performance, transfers to broader VQA and text-only QA benchmarks, and achieves lower inference latency than modular MM-RAG pipelines. Together, these results suggest that retrieval-augmented multimodal reasoning should not only ask \emph{what} knowledge to retrieve, but also \emph{where} in the image that knowledge should be grounded.

\paragraph{Future works.}
MM-RAG alone does not resolve visual perception. By aligning hidden states with pixel masks and retrieving evidence from crops, \textsc{PixSearch} partly bypasses long-tail entity recognition, but incorrect localization and missed visual attributes remain bottlenecks. Mixture of Layers~\citep{kim2026mixturelayers} and DeepStack~\citep{meng2024deepstack} motivate complementary improvements through instruction-conditioned visual-layer routing and deeper visual-token integration, respectively. Future work should jointly improve grounded perception, mask alignment, and retrieval: external facts cannot replace observing image-specific attributes or spatial relations. Evaluations that hold external knowledge fixed while varying these visual properties could test whether models actually use the relevant visual evidence.

\FloatBarrier
\section*{Limitations}
\textsc{PixSearch} depends on accurate region grounding and the coverage and quality of retrieved evidence; segmentation errors or missing knowledge can still lead to incorrect answers. GPT-4.1-generated supervision and LLM-based evaluation may introduce biases and errors. Our benchmark results do not establish robustness across languages or real-world wearable environments, and lower inference latency than modular pipelines does not establish feasibility for on-device deployment.

\clearpage
\bibliographystyle{assets/plainnat}
\bibliography{paper}
\clearpage
\beginappendix
\suppressfloats[t]
\section{Generalization to Text-Only QA}
\label{app:text_qa}
We evaluate HotpotQA~\citep{yang2018hotpotqa}, NQ~\citep{kwiatkowski-etal-2019-nq, joshi2017triviaqa}, PopQA~\citep{mallen2023popqa}, and MuSiQue~\citep{trivedi2022musique}, reporting relaxed exact match (EM) and F1. For \textsc{PixSearch} variants, results are means and standard deviations over three independent training seeds.

Despite not being trained on text-only QA data, \textsc{PixSearch} also performs strongly on out-of-domain textual QA benchmarks spanning multi-hop, open-domain, and entity-centric reasoning (Table~\ref{tab:text_only_qa_em_f1}), demonstrating preserved textual reasoning ability.

\begin{table}[!htbp]
\centering
\begin{minipage}{.90\textwidth}
\centering
\small
\setlength{\tabcolsep}{6pt}
\renewcommand{\arraystretch}{1.1}
\begin{adjustbox}{max width=\columnwidth}
\begin{tabular}{@{}lcccc@{}}
\toprule
\textbf{Model} & \textbf{HotpotQA} & \textbf{NQ} & \textbf{PopQA} & \textbf{MuSiQue} \\
\midrule
LLaVA & \makecell{8.33\\10.25} & \makecell{10.18\\11.06} & \makecell{4.70\\7.98} & \makecell{5.70\\6.35} \\
PLUM & \makecell{15.30\\18.80} & \makecell{15.00\\18.35} & \makecell{7.40\\9.58} & \makecell{6.60\\8.43} \\
\makecell[l]{PixSearch\\\textit{No Search}} & \makecell{$23.7\pm 1.2$\\$27.1\pm 1.0$} & \makecell{$30.3\pm 1.0$\\$33.2\pm 1.1$} & \makecell{$17.4\pm 0.8$\\$17.9\pm 0.8$} & \makecell{$10.0\pm 0.6$\\$12.9\pm 0.7$} \\
\makecell[l]{\textbf{PixSearch}\\\textit{Full}} & \makecell{$\textbf{26.5}\pm 1.0$\\$\textbf{28.2}\pm 1.3$} & \makecell{$\textbf{31.2}\pm 1.1$\\$\textbf{33.7}\pm 1.2$} & \makecell{$\textbf{31.9}\pm 1.2$\\$\textbf{32.6}\pm 1.2$} & \makecell{$\textbf{11.7}\pm 0.7$\\$\textbf{13.0}\pm 0.7$} \\
\bottomrule
\end{tabular}
\end{adjustbox}
\caption{\textbf{Text-only QA performance evaluation.} Scores are EM/F1 in \%. All models have 13B parameters. For PixSearch variants, we report mean $\pm$ std over three independent training seeds. Each cell shows EM (top) and F1 (bottom).}
\label{tab:text_only_qa_em_f1}
\end{minipage}
\end{table}

\section{Hyperparameters and Compute Details}
In Table \ref{tab:hyperparameters}, we detail the hyperparameter settings for \textsc{PixSearch} and our backbone model, PLUM~\citep{blume-kim-2025partonomy}.

\begin{table}[!htbp]
\centering
\begin{minipage}{.95\textwidth}
\centering
\small
\setlength{\tabcolsep}{6pt}
\begin{tabularx}{\columnwidth}{@{}YYY@{}}
\toprule
\textbf{Hyperparameter} & \textbf{PLUM} & \textbf{PixSearch} \\
\midrule
\multicolumn{3}{l}{\textit{Backbone}}\\
Language model & LLaVA-13B & LLaVA-13B (PLUM init.) \\
Vision tower & CLIP ViT-L/14 & CLIP ViT-L/14 \\
Mask decoder & SAM ViT-H & SAM ViT-H \\
\midrule
\multicolumn{3}{l}{\textit{Training schedule}}\\
Input resolution & $1024^2$ & $1024^2$ \\
Max text length & 512 & 512 \\
Precision & bf16 & bf16 \\
Epochs & 25 + 4 & Stage-1: 20, Stage-2: 20 \\
Batch size & 6 & 6 \\
Grad. accumulation & 10 & 10 \\
\midrule
\multicolumn{3}{l}{\textit{Optimizer}}\\
Optimizer & AdamW & AdamW \\
LR & $3\!\times\!10^{-4}$ & $2\!\times\!10^{-4}$ (S1), $3\!\times\!10^{-4}$ (S2) \\
Betas & (0.9, 0.95) & (0.9, 0.95) \\
Weight decay & 0 & 0 \\
\midrule
\multicolumn{3}{l}{\textit{Loss weights}}\\
$\lambda_{\mathrm{CE}}$ & 1.0 & 1.0 \\
$\lambda_{\mathrm{seg}}$ & 8.0 & 8.0 \\
$\lambda_{\mathrm{BCE}}$ & 2.0 & 2.0 \\
$\lambda_{\mathrm{KL}}$ & 0.1 & 0.1 \\
$\lambda_{\mathrm{cls}}$ & 2.0 & 2.0 \\
\midrule
\multicolumn{3}{l}{\textit{Modules}}\\
BIO span tagger & \cmark & \cmark \\
Bidirectional encoder & 2048 & 2048 \\
Feedback Loop & \cmark & \cmark \\
Trainable SAM components & decoder+prompt enc. & decoder+prompt enc. \\
LoRA on LM (q,v) & $r=8$ & $r=8$ \\
\bottomrule
\end{tabularx}
\caption{\textbf{Hyperparameters for PLUM and PixSearch.} Stage-2 settings follow the documented completed run.}
\label{tab:hyperparameters}
\end{minipage}
\end{table}

\paragraph{Stage-2 training and compute.}
The documented completed fine-tuning run used eight NVIDIA H100 80GB GPUs with bf16 precision, a per-GPU batch size of 6, and 10 gradient-accumulation steps (effective batch size 480). It ran for 20 epochs of 500 optimizer steps, totaling 10,000 updates, with learning rate $3\times10^{-4}$, maximum text length 512, and random seed 42. Semantic segmentation, referring-expression segmentation, visual instruction tuning, reasoning segmentation, and search-interleaved data were sampled in the ratio $3{:}3{:}5{:}1{:}9$. These epochs denote fixed update budgets rather than exhaustive passes through each dataset.

The recorded elapsed runtime was approximately 61.5 hours (492 GPU-hours); summing the logged training-step durations gives 56.0 hours (448 GPU-hours), excluding time outside the training loop. These figures describe one completed Stage-2 run, not the total experimental budget: they exclude backbone training, other runs and ablations, and synthetic-data generation. Complete Stage-1 timing was not available in the retained logs.

\paragraph{Training software.}
The recorded environment used Python 3.10.13, PyTorch 2.0.1, Transformers 4.31.0, and DeepSpeed 0.9.5.

\section{Decode-with-Retrieval Algorithm Walkthrough}
\label{sec:detailed_decode}

Algorithm~\ref{alg:decode_with_retrieval} describes the search--interleaved
decoding mechanism used by PixSearch. Below is a detailed walkthrough.

\paragraph{Autoregressive decoding with retrieval control.}
At each decoding step, the model autoregressively predicts the next token. If the
token is a normal language token, decoding continues normally. When the model
emits the special token \texttt{<search>}, it signals that retrieval is needed.

\paragraph{Stack-based parsing of retrieval spans.}
Each retrieval request is enclosed within \texttt{<search>} ... \texttt{</search>}.
To correctly pair them (especially when multiple retrieval calls occur in a
single answer), a stack is maintained. The index of each \texttt{<search>} token
is pushed on the stack; when the model later emits \texttt{</search>}, that
interval defines a payload containing the retrieval modality and/or textual
query.

\paragraph{Determining retrieval modality.}
Inside the \texttt{<search>} block, the model emits one of:
\begin{itemize}
\item \textbf{\texttt{<image>}}: retrieve using the entire image.
\item \textbf{\texttt{<region>}}: call the mask decoder to predict a segmentation
mask for the referred entity; crop the image using the mask.
\item \textbf{\texttt{<text>}}: use the generated textual span as a query.
\end{itemize}

\paragraph{Executing retrieval.}
The system then calls \texttt{search\_api(query, k)}, where \(k\) is the number
of returned documents. Retrieved snippets are formatted and injected back into
the generation sequence using \texttt{<information>} ... \texttt{</information>}.

\paragraph{Search-interleaved reasoning.}
PixSearch may perform multiple retrievals throughout a single answer. Retrieved
evidence stays in the context, enabling multihop reasoning grounded in both the
image and external knowledge.

\paragraph{Termination.}
Decoding continues until an end-of-sequence token is reached.

\begin{algorithm}[!htbp]
\small
\caption{Decode with Retrieval}
\label{alg:decode_with_retrieval}
\begin{algorithmic}[1]
\Require
  \Statex $M$: multimodal model, $I$: input image
  \Statex $\texttt{search\_api}(q,k)$: retrieval function
\Ensure
  \Statex Generated sequence augmented with retrieved info

\State $\textit{gen} \gets \texttt{prompt\_with\_image}(I)$, \quad $\textit{stack} \gets [\;]$
\While{not EOS and steps remaining}
  \State $\textit{tok} \gets M.\texttt{generate\_next}(\textit{gen})$; \quad $\textit{gen} \gets \textit{gen} + \textit{tok}$
  \If{ends with ``\texttt{<search>}''} push index onto $\textit{stack}$
  \ElsIf{ends with ``\texttt{</search>}'' and $\textit{stack}$ not empty}
     \State $\textit{payload} \gets \texttt{slice}(\textit{gen}, \texttt{pop}(\textit{stack}))$
     \State $\textit{mode} \gets \texttt{parse\_payload}(\textit{payload})$
     \If{$\textit{mode}$ = ``\texttt{<region>}''}
        \State $\textit{mask} \gets \texttt{predict\_mask}(\texttt{last\_entity}(\textit{gen}), I)$; \; $\textit{query} \gets \texttt{crop}(I,\textit{mask})$
     \ElsIf{$\textit{mode}$ = ``\texttt{<img>}''} \State $\textit{query} \gets I$
     \Else \State $\textit{query} \gets \textit{payload\_text}$
     \EndIf
     \State $\textit{info} \gets \texttt{format}(\texttt{search\_api}(\textit{query},k))$
     \State $\textit{gen} \gets \texttt{append}(\textit{gen}, \texttt{<information>} + \textit{info} + \texttt{</information>})$
  \EndIf
\EndWhile
\State \Return $\textit{gen}$
\end{algorithmic}
\end{algorithm}

\section{Explanation of the Ten Visual Question Types}
\label{app:question_types}

Table~\ref{tab:question_types} lists the ten question categories used for
Stage-2 SFT construction. Below we provide expanded definitions.

\paragraph{OCR Read.}
Questions requiring verbatim transcription of scene text.

\paragraph{OCR + Visual Reasoning.}
Requires interpreting the meaning of text in context (e.g., a scoreboard).

\paragraph{Multi-hop External Knowledge Reasoning.}
Requires chaining two or more knowledge lookups (e.g., identify entity \(\rightarrow\)
retrieve its founding date).

\paragraph{Fine-grained Entity Identification.}
Entity-level identification often requiring region-level cropping.

\paragraph{Visual Reasoning -- Attribute.}
Recognition of visible attributes (color, shape, texture).

\paragraph{Visual Reasoning -- Counting.}
Counting objects in the image.

\paragraph{Visual Reasoning -- Binary.}
Yes/no visual questions.

\paragraph{Social Commonsense Reasoning.}
Inferring human motivations or social context.

\paragraph{Physical Commonsense Reasoning.}
Inferring physical affordances, constraints, and outcomes.

\paragraph{Factoid / KB Questions.}
Open-domain factual questions referencing entities in or implied by the image.

\begin{table}[!htbp]
\small
\centering
\begin{minipage}{.92\textwidth}
\centering

\begin{tabularx}{\columnwidth}{@{}YY@{}}
\toprule
\textbf{Question Type} & \textbf{Example Question} \\
\midrule
\textbf{OCR Read} & ``What does it say near the tail of the plane?'' \\
\textbf{OCR + Visual Reasoning} & ``Which team is winning the game?'' \\
\textbf{Multi-hop External Knowledge Reasoning} & ``When was the soft-drink company shown first created?'' \\
\textbf{Fine-grained Entity Identification} & ``What class of animal is this creature?'' \\
\textbf{Visual Reasoning -- Attribute} & ``What is the color of the car in the background?'' \\
\textbf{Visual Reasoning -- Counting} & ``How many cars are there in the image?'' \\
\textbf{Visual Reasoning -- Binary} & ``Is the man in the image wearing a hat?'' \\
\textbf{Social Commonsense Reasoning} & ``Why might the seated man have trouble getting around?'' \\
\textbf{Physical Commonsense Reasoning} & ``What could block the washer's door?'' \\
\textbf{Factoid / KB Questions} & ``Hot dogs were invented in which country?'' \\
\bottomrule
\end{tabularx}
\caption{\textbf{Overview of the ten visual question types used in Stage-2.}}
\label{tab:question_types}
\end{minipage}
\end{table}

\section{In-Context Learning Samples}

All examples are displayed in Tables \ref{tab:question_selection}, \ref{tab:question_decomposition}, \ref{tab:response_generation}, \ref{tab:query_type_assignment}

\subsection{Question Selection ICL Examples}

To teach the model \emph{when} retrieval is needed, we constructed Question Selection
examples that expose the model to a diverse set of visual questions drawn from
CRAG-MM, OK-VQA, and InfoSeek. Each example pairs a question with an image and a
binary label indicating whether external knowledge is required. The key design
principle is that retrieval is only beneficial when the image alone cannot
resolve the question. 

Thus, the examples include:
(1) fine-grained or long-tail entity identification tasks (e.g., identifying car
models, drink brands, or rare animals), which require region-level or whole-image
search;
(2) multi-step factual or encyclopedic queries (e.g., historical dates, object
origins), where knowledge beyond the image is essential;
and (3) questions solvable purely from visual inspection (e.g., ``Translate this'',
``What is the couch made of?'', ``What grade is the child in?''), where retrieval
would be unnecessary or potentially harmful.

By contrasting retrieval and no-retrieval cases with high visual similarity, the
model learns a robust policy for deciding \textit{when} to trigger
\texttt{<search>} calls.

\begin{table}[!htbp]
\centering
\small

\setlength{\tabcolsep}{3pt}
\begin{tabularx}{\textwidth}{@{}p{3.3cm}Yp{1.3cm}@{}}
\toprule
\textbf{Image File} & \textbf{Question} & \textbf{Retrieval} \\
\midrule

\nolinkurl{cragmm/4ec6f8ae.png} & How many hybrid variations of this car were there in 2024? & Yes \\

\nolinkurl{cragmm/08629717.png} & Is that drink good for my gut health? & Yes \\

\nolinkurl{cragmm/fb2fed47.png} & How many arms does this statue typically have? & Yes \\

\nolinkurl{cragmm/1c613a06.png} & Translate this. & No \\

\nolinkurl{cragmm/569a3617.png} & Which station has more tracks, this one or Penn Station? & Yes \\

\nolinkurl{cragmm/a97e2470.png} & Where was the designer who developed this car originally from? & Yes \\

\nolinkurl{cragmm/4f81b083.png} & What is the seating capacity of the car with the open trunk? & Yes \\

\nolinkurl{cragmm/b797333f.png} & What does the word ``skrzela'' translate to in English? & No \\

\nolinkurl{okvqa/3575845.png} & What is the couch made of? & No \\

\nolinkurl{cragmm/d253fc27.png} & In what year did the president for whom this bridge is named win the Battle of Trenton? & Yes \\

\nolinkurl{okvqa/4597935.png} & Where is the farm depicted on the sign located? & Yes \\

\nolinkurl{okvqa/3742825.png} & What brand of car is this? & Yes \\

\nolinkurl{okvqa/3182455.png} & The fabric on that couch was very popular in the eighties --- what was it called? & Yes \\

\nolinkurl{okvqa/1981195.png} & Why might the man be kicking up sand? & No \\

\nolinkurl{okvqa/1217825.png} & What holiday might they be celebrating? & Yes \\

\nolinkurl{okvqa/3778685.png} & With what religious tradition is the creature portrayed here associated? & Yes \\

\bottomrule
\end{tabularx}
\caption{\textbf{Question Selection Examples.} We show representative questions, whether retrieval was needed or not, and the associated image filenames. The \textbf{Image File} paths were truncated to have only the prefix of the image path for brevity.}
\label{tab:question_selection}
\end{table}

\subsection{Question Decomposition ICL Examples}

Question Decomposition examples teach the model to break down complex questions
into a sequence of \emph{atomic}, retrieval-ready sub-questions. The rationale is
that many visual knowledge queries involve implicit multi-hop reasoning (e.g.,
identify the entity in the image, then query its properties). To capture this,
the examples label questions as either decomposable or non-decomposable, and
provide the exact sub-questions that should be produced.

Decomposable cases typically involve:
(1) entity grounding followed by factual lookup (e.g., identify the king of
Spain \(\rightarrow\) find when he became king);
(2) place or object recognition followed by knowledge retrieval (e.g., identify
the arena \(\rightarrow\) retrieve capacity);
or (3) multi-hop knowledge chains (e.g., identify the farm \(\rightarrow\) locate it).

Non-decomposable examples demonstrate when a single visual or commonsense step
suffices (e.g., ``Translate this'', ``What kind of sculpture is this?''). This
contrastive supervision helps the model learn when multi-hop decomposition is
beneficial and when it is unnecessary.

\begin{table}[!htbp]
\centering
\small

\setlength{\tabcolsep}{3pt}
\begin{tabularx}{\textwidth}{@{}p{3.3cm}YY@{}}
\toprule
\textbf{Image File} & \textbf{Original Question} & \textbf{Sub-Questions} \\
\midrule

\nolinkurl{cragmm/4dcc84dc.png} &
When did the king of that country become king? &
1) Who is the king of Spain? \newline
2) When did the king of Spain become king? \\

\nolinkurl{cragmm/da33192e.png} &
What's the capacity of this arena? &
1) What is this place? \newline
2) What's the capacity of this arena? \\

\nolinkurl{cragmm/f73ab93c.png} &
Where was the first sign accompanying this erected? &
1) Where was the first pedestrian crossing signal erected? \\

\nolinkurl{cragmm/878088c7.png} &
What's the ideal temperature for this plant? &
1) What is this plant? \newline
2) What is the ideal temperature for this plant? \\

\nolinkurl{cragmm/b03b7dd6.png} &
How long can I use it without turning it off? &
1) What is the model name of this generator? \newline
2) How long can I use this generator? \\

\nolinkurl{cragmm/ec87776d.png} &
Translate this into English. &
1) Translate this into English. \\

\nolinkurl{cragmm/c0a60302.png} &
What kind of sculpture is this? &
1) What kind of sculpture is this? \\

\nolinkurl{cragmm/d253fc27.png} &
In what year did the president for whom this bridge is named win the Battle of Trenton? &
1) What is the name of the bridge? \newline
2) What year did the president win the Battle of Trenton? \\

\nolinkurl{okvqa/804725.png} &
What grade is the child in? &
1) What grade is the child in? \\

\nolinkurl{okvqa/3575845.png} &
What is the couch made of? &
1) What is the model name of this couch? \\

\nolinkurl{okvqa/4830705.png} &
Which historical group wore that clothing accessory? &
1) Which historical group wore spurs as a clothing accessory? \\

\nolinkurl{okvqa/4597935.png} &
Where is the farm depicted on the sign located? &
1) What is the farm name? \newline
2) Where is the farm located? \\

\nolinkurl{okvqa/5246795.png} &
What is the purpose of the purple vehicle depicted? &
1) What is the purpose of the purple vehicle depicted? \\

\nolinkurl{okvqa/3742825.png} &
What brand of car is this? &
1) What brand of car is this? \\

\nolinkurl{infoseek/04308592.JPEG} &
What country does this drink belong to? &
1) Where was Louis Jadot made in? \\

\nolinkurl{infoseek/04123114.JPEG} &
What is the basionym of this plant? &
1) What is this plant? \newline
2) What is the basionym of this plant? \\

\bottomrule
\end{tabularx}
\caption{\textbf{Question Decomposition Examples.} Each example lists the original question, whether it was decomposed, and the resulting sub-questions.}
\label{tab:question_decomposition}
\end{table}

\begin{table}[!htbp]
\centering
\small

\setlength{\tabcolsep}{3pt}
\begin{tabularx}{\textwidth}{@{}p{3.3cm}>{\hsize=1.4\hsize}Y>{\hsize=.6\hsize}Y@{}}
\toprule
\textbf{Image File} & \textbf{Question + Model Answer} & \textbf{Ground-Truth Answer} \\
\midrule

\nolinkurl{cragmm/4dcc84dc.png} &
Q: When did the king of that country become king? \newline
A: The paper says ``\ldots{} independence from Spain.'' Since Mexico has no king, this refers to Spain. The king of Spain \texttt{<search>} is King Felipe VI. Felipe VI came to the throne \texttt{<search>} on June 19, 2014. &
June 19, 2014 \\

\nolinkurl{cragmm/da33192e.png} &
Q: What's the capacity of this arena? \newline
A: The place shown \texttt{<search>} is Madison Square Garden. It can house \texttt{<search>} between 18,006 and 20,789 people. &
18,006--20,789 depending on event \\

\nolinkurl{cragmm/f73ab93c.png} &
Q: Where was the first sign accompanying this erected? \newline
A: The first pedestrian crossing signal was erected in \texttt{<search>} Bridge Street, Westminster, London, in December 1868. &
Bridge Street, Westminster, London (1868) \\

\nolinkurl{cragmm/878088c7.png} &
Q: What's the ideal temperature for this plant? \newline
A: The plant \texttt{<search>} appears to be Ravenea rivularis. Ravenea rivularis prefers \texttt{<search>} 65--85\textdegree{}F. &
Prefers 65--85\textdegree{}F \\

\nolinkurl{cragmm/b03b7dd6.png} &
Q: How long can I use it without turning it off? \newline
A: The generator appears to be \texttt{<search>} a Firman Tri Fuel Generator. Its usage time is \texttt{<search>} a maximum of 12 hours with a 2--3 hour recommended break. &
12 hours, with 2--3 hour breaks \\

\nolinkurl{cragmm/ec87776d.png} &
Q: Translate this into English. \newline
A: It translates to: ``What mixer is used in a Paloma?'' &
``What mixer is used in a Paloma?'' \\

\nolinkurl{cragmm/c0a60302.png} &
Q: What kind of sculpture is this? \newline
A: This is a snowman with red hat and carrot nose. &
Snowman balloon figure with red hat and carrot nose \\

\nolinkurl{cragmm/d253fc27.png} &
Q: In what year did the president for whom this bridge is named win the Battle of Trenton? \newline
A: The bridge \texttt{<search>} is the George Washington Bridge. George Washington won the Battle of Trenton \texttt{<search>} on December 26, 1776. &
1776 \\

\nolinkurl{okvqa/804725.png} &
Q: What grade is the child in? \newline
A: The child appears around 5 years old, which corresponds to kindergarten. &
Kindergarten \\

\nolinkurl{okvqa/3575845.png} &
Q: What is the couch made of? \newline
A: The couch \texttt{<search>} appears to be a sofa made of fabric, cotton, wood, foam, and leather. &
Fabric / cotton / wood / foam \\

\nolinkurl{okvqa/4830705.png} &
Q: Which historical group wore that clothing accessory? \newline
A: Spurs were worn by \texttt{<search>} cowboys. &
Cowboys \\

\nolinkurl{okvqa/4597935.png} &
Q: Where is the farm depicted on the sign located? \newline
A: The farm \texttt{<search>} is Victoria Garden Centre Farm, located \texttt{<search>} in the Midwest US. &
Midwest US \\

\nolinkurl{okvqa/5246795.png} &
Q: What is the purpose of the purple vehicle depicted? \newline
A: It is a tourist bus used to transport people. &
Tourist transport / mass transit \\

\nolinkurl{okvqa/3742825.png} &
Q: What brand of car is this? \newline
A: The brand \texttt{<search>} is Volkswagen. &
Volkswagen \\

\nolinkurl{infoseek/04308592.JPEG} &
Q: What country does this drink belong to? \newline
A: The drink is Louis Jadot, which was made in \texttt{<search>} France. &
France \\

\nolinkurl{infoseek/04123114.JPEG} &
Q: What is the basionym of this plant? \newline
A: The plant \texttt{<search>} appears to be a type of tatsoi mustard. Its basionym is \texttt{<search>} Cupressus lawsoniana. &
Brassica rapa / Brassica rapa var. rosularis \\

\bottomrule
\end{tabularx}
\caption{\textbf{Response Generation Examples.} Each example includes the question, the model-generated answer (which includes \texttt{<search>} calls), and the ground-truth answer.}
\label{tab:response_generation}
\end{table}

\subsection{Response Generation ICL Examples}

Response Generation samples demonstrate how the model should integrate external
knowledge into coherent, grounded answers by interleaving \texttt{<search>}
tokens with natural language output. Each example provides:
(1) the question,
(2) the model's expected search-interleaved reasoning trajectory with properly
placed \texttt{<search>} and \texttt{</search>} markers, and
(3) the corresponding ground-truth answer.

The central rationale is to teach the model not only to ask for external
knowledge, but to do so at the correct semantic point within the reasoning
process. For instance, the model must first identify ``King Felipe VI'' before
issuing a second retrieval about his coronation date. 

These examples also illustrate how retrieved facts are integrated back into the
narrative, enabling PixSearch to produce faithful, factual answers without
hallucinating details or over-triggering retrieval. The contrast between the
final grounded answer and the intermediate reasoning highlights how to combine
multiple \texttt{<search>} calls into a single, well-structured response.

\begin{table}[!htbp]
\centering
\fontsize{8}{9.5}\selectfont

\setlength{\tabcolsep}{3pt}
\begin{tabularx}{\textwidth}{@{}p{3.3cm}>{\hsize=1.3\hsize}Yp{1.25cm}>{\hsize=.7\hsize}Y@{}}
\toprule
\textbf{Image File} &
\textbf{Question + \texttt{<search>}-Interleaved Response} &
\textbf{Query Types} &
\textbf{Queries} \\
\midrule

\nolinkurl{cragmm/4dcc84dc.png} &
Q: When did the king of that country become king? \newline
A: The king of Spain \texttt{<search>} is King Felipe VI. \texttt{</search>}%
He came to the throne \texttt{<search>} on June 19, 2014. \texttt{</search>} &
TEXT, TEXT &
1) Who is the king of Spain? \newline 2) When did King Felipe VI become king? \\

\nolinkurl{cragmm/da33192e.png} &
Q: What's the capacity of this arena? \newline
A: The place shown \texttt{<search>} is Madison Square Garden. \texttt{</search>}%
It can house \texttt{<search>} between 18{,}006 and 20{,}789 people. \texttt{</search>} &
IMAGE, TEXT &
1) image \newline 2) What is the capacity of this arena? \\

\nolinkurl{cragmm/f73ab93c.png} &
Q: Where was the first sign accompanying this erected? \newline
A: The first pedestrian crossing signal was erected \texttt{<search>} in Bridge Street, Westminster, London. \texttt{</search>} &
TEXT &
Where was the first pedestrian crossing signal erected? \\

\nolinkurl{cragmm/878088c7.png} &
Q: What's the ideal temperature for this plant? \newline
A: The plant \texttt{<search>} appears to be Ravenea rivularis. \texttt{</search>}%
It prefers \texttt{<search>} 65--85\textdegree{}F. \texttt{</search>} &
REGION, TEXT &
1) plant \newline 2) Ideal temperature for Ravenea rivularis \\

\nolinkurl{cragmm/b03b7dd6.png} &
Q: How long can I use it without turning it off? \newline
A: The generator \texttt{<search>} appears to be a Firman Tri Fuel Generator. \texttt{</search>}%
Its usage time is \texttt{<search>} a maximum of 12 hours with breaks. \texttt{</search>} &
REGION, TEXT &
1) generator \newline 2) Use time for Firman Tri Fuel Generator \\

\nolinkurl{cragmm/ec87776d.png} &
Q: Translate this into English. \newline
A: The sentence translates to: ``What mixer is used in a Paloma?'' &
--- &
--- \\

\nolinkurl{cragmm/c0a60302.png} &
Q: What kind of sculpture is this? \newline
A: This is a snowman with red hat and carrot nose. &
--- &
--- \\

\nolinkurl{cragmm/d253fc27.png} &
Q: In what year did the president for whom this bridge is named win the Battle of Trenton? \newline
A: The bridge \texttt{<search>} is the George Washington Bridge. \texttt{</search>}%
Washington won the battle \texttt{<search>} on December 26, 1776. \texttt{</search>} &
REGION, TEXT &
1) bridge \newline 2) Battle of Trenton date \\

\nolinkurl{okvqa/804725.png} &
Q: What grade is the child in? \newline
A: The child appears around 5 years old, corresponding to kindergarten. &
--- &
--- \\

\nolinkurl{okvqa/3575845.png} &
Q: What is the couch made of? \newline
A: The couch \texttt{<search>} appears to be a sofa made of fabric, cotton, wood, foam, and leather. \texttt{</search>} &
REGION &
couch \\

\nolinkurl{okvqa/4830705.png} &
Q: Which historical group wore that clothing accessory? \newline
A: Spurs were commonly worn by \texttt{<search>} cowboys. \texttt{</search>} &
TEXT &
What historical group used to wear spurs? \\

\nolinkurl{okvqa/4597935.png} &
Q: Where is the farm depicted on the sign located? \newline
A: The farm \texttt{<search>} is Victoria Garden Centre Farm. \texttt{</search>}%
It is located \texttt{<search>} in the Midwest US. \texttt{</search>} &
IMAGE, TEXT &
1) image \newline 2) Where is Victoria Garden Centre Farm located? \\

\nolinkurl{okvqa/5246795.png} &
Q: What is the purpose of the purple vehicle depicted? \newline
A: It is a tourist bus used to transport people for sightseeing. &
--- &
--- \\

\nolinkurl{okvqa/3742825.png} &
Q: What brand of car is this? \newline
A: The car \texttt{<search>} is Volkswagen. \texttt{</search>} &
REGION &
car \\

\nolinkurl{infoseek/04308592.JPEG} &
Q: What country does this drink belong to? \newline
A: The drink is Louis Jadot, made in \texttt{<search>} France. \texttt{</search>} &
TEXT &
Where was Louis Jadot made? \\

\nolinkurl{infoseek/04123114.JPEG} &
Q: What is the basionym of this plant? \newline
A: The plant \texttt{<search>} appears to be tatsoi mustard. \texttt{</search>}%
Its basionym is \texttt{<search>} Cupressus lawsoniana. \texttt{</search>} &
REGION, TEXT &
1) plant \newline 2) Basionym of tatsoi mustard \\

\bottomrule
\end{tabularx}
\caption{\textbf{Query-Type Assignment Examples.} Each example includes the original question, the model's \texttt{<search>}-interleaved response, and the corresponding query types and queries.}
\label{tab:query_type_assignment}
\end{table}

\subsection{Query-Type Assignment ICL Examples}

To teach the model \emph{how} to choose the correct retrieval modality,
Query-Type Assignment examples pair each search-interleaved response with:
(1) the modality chosen for each \texttt{<search>} call (TEXT, IMAGE, or REGION),
and (2) the exact query string or region reference used.

The rationale behind these examples is rooted in the observation that different
question types require different forms of evidence:
\begin{itemize}[nosep,leftmargin=*]
\item \textbf{REGION} queries are necessary when a specific entity must be
identified or disambiguated (e.g., car brand, plant type, or farm name).
\item \textbf{IMAGE} queries are appropriate when holistic scene recognition is
required (e.g., identifying Madison Square Garden from a full stadium view).
\item \textbf{TEXT} queries are used for abstract facts requiring no visual input
(e.g., ``When did Washington win the Battle of Trenton?'').
\end{itemize}

By aligning each retrieval call with its intended modality, these examples teach
the model to properly route retrieval operations and to generate the correct
query based on the context of the ongoing reasoning process. This ensures that
PixSearch issues retrieval in a controlled, modality-aware manner that improves
factual precision and reduces retrieval noise.

\section{Training Controls}
\label{app:training_controls}
Table~\ref{tab:training_controls_full} expands the training comparison in Table~\ref{tab:stage1_ablation}. The controls compare Stage-2-only SFT, single-stage joint SFT, two-stage SFT, and Search-R1-style reinforcement learning using GRPO~\citep{jin2025searchr1}, under matched data and optimization budgets. Matching optimization budgets does not imply equal GPU-hours, since RL incurs rollout-generation costs. The compute measurement reported in the hyperparameter appendix covers one completed two-stage model's Stage-2 SFT run; corresponding compute measurements for the other controls are not available.

\begin{table}[!ht]
\centering
\begin{minipage}{.82\textwidth}
\centering
\small
\setlength{\tabcolsep}{6pt}
\begin{tabular}{@{}lccccc@{}}
\toprule
Training & Acc. & R@1 & R@10 & gIoU & mIoU \\
\midrule
Stage-2 SFT only & 18.5 & 0.05 & 0.19 & 14.8 & 10.2 \\
Joint SFT & 32.6 & 0.18 & 0.35 & 67.4 & 46.8 \\
Two-stage SFT & \textbf{36.9} & \textbf{0.22} & \textbf{0.41} & \textbf{78.9} & \textbf{55.98} \\
Search-R1-style GRPO & \textbf{36.9} & 0.20 & 0.38 & 71.6 & 50.3 \\
\bottomrule
\end{tabular}
\caption{\textbf{Expanded training comparison.} Accuracy is on CRAG-MM, gIoU on RefCOCO, and mIoU on ADE20K; these three metrics are percentages. Retrieval recall is reported on a $[0,1]$ scale.}
\label{tab:training_controls_full}
\end{minipage}
\end{table}

Two-stage SFT exceeds joint SFT by 4.3 accuracy points and 11.5 RefCOCO gIoU points. GRPO matches its answering accuracy, but two-stage SFT has higher RefCOCO gIoU by 7.3 points and ADE20K mIoU by 5.68 points. These results support better grounding retention for the tested SFT schedule, rather than a general claim that SFT outperforms RL or that joint optimization necessarily collapses.

\FloatBarrier
\Needspace{30\baselineskip}
\section{Inference Latency Breakdown}
\label{app:latency_breakdown}
Table~\ref{tab:latency_components} decomposes the end-to-end latency in Table~\ref{tab:latency} into model inference, mask prediction and cropping, query encoding, and vector-database retrieval. Retrieval time is included in the total. Call counts are the average number of retrieval calls per example. The cached GroundedSAM--LLaVA pipeline excludes online GPT-4.1 entity extraction; the GPT-assisted variant includes this external step.

\begin{table}[!ht]
\centering
\begin{minipage}{.76\textwidth}
\centering
\small
\setlength{\tabcolsep}{6pt}
\begin{tabular}{@{}lcccc@{}}
\toprule
Component & Full & Whole & Cached & +GPT \\
\midrule
Model inference (s) & 1.76 & 1.72 & 1.84 & 2.44 \\
Mask/crop (s) & 0.23 & --- & 1.60 & 1.60 \\
Query encoding (s) & 0.12 & 0.11 & 0.13 & 0.13 \\
Retrieval (s) & 0.69 & 0.62 & 0.83 & 0.83 \\
\midrule
Total (s) & 2.80 & 2.45 & 4.40 & 5.00 \\
p90 (s) & 3.30 & 2.96 & 4.10 & 5.90 \\
Mean calls & 3.0 & 2.4 & 2.9 & 2.9 \\
\bottomrule
\end{tabular}
\caption{\textbf{Latency components on CRAG-MM.} Full and Whole denote PixSearch with full and whole-image retrieval, respectively. Cached and +GPT denote the cached and GPT-4.1-assisted GroundedSAM--LLaVA pipelines.}
\label{tab:latency_components}
\end{minipage}
\end{table}

Full PixSearch is 0.35\,s slower than whole-image retrieval, but reduces latency by 36.4\% relative to the cached pipeline and 44\% relative to the GPT-assisted pipeline. Its mask/crop component takes 0.23\,s, compared with 1.60\,s for the separate grounding pipeline. The primary latency comparison therefore uses the cached baseline, so its advantage does not depend on the online entity-extraction call.

Region localization operates on the query image: the predicted mask defines a crop that is encoded for retrieval against the image index. It does not require indexing every possible region or mask in the retrieval corpus.

\end{document}